\documentclass[runningheads]{llncs}
\usepackage{graphicx}
\usepackage{comment}
\usepackage{amsmath,amssymb} % define this before the line numbering.
\usepackage{color}
\usepackage{subfigure}
\usepackage[linesnumbered,boxed,ruled,commentsnumbered]{algorithm2e}
\usepackage{verbatim}
\usepackage{multirow}

\begin{document}
% \linenumbers
\pagestyle{headings}
\mainmatter
\def\ECCVSubNumber{2598}  % Insert your submission number here

\title{Image Stitching Based on Planar Region Consensus} % Replace with your title

\begin{comment}
\titlerunning{ECCV-20 submission ID \ECCVSubNumber}
\authorrunning{ECCV-20 submission ID \ECCVSubNumber}
\author{Anonymous ECCV submission}
\institute{Paper ID \ECCVSubNumber}
\end{comment}
%******************

% \begin{comment}
\titlerunning{Image Stitching Based on Planar Region Consensus}
\author{Aocheng Li\and
Jie Guo\and
Yanwen Guo}
\authorrunning{Li et al.}
\institute{Nanjing University, Department of Computer Science and Technology
\email{\{161220062@smail,jieguo,ywguo\}@nju.edu.cn}}
% \end{comment}
%******************
\maketitle

\begin{abstract}
Image stitching for two images without a global transformation between them is notoriously difficult. In this paper, noticing the importance of planar structure under perspective geometry, we propose a new image stitching method which stitches images by allowing for the alignment of a set of matched dominant planar regions. Clearly different from previous methods resorting to plane segmentation, the key to our approach is to utilize rich semantic information directly from RGB images to extract planar image regions with a deep Convolutional Neural Network (CNN). We specifically design a new module to make fully use of existing semantic segmentation networks to accommodate planar segmentation. To train the network, a dataset for planar region segmentation is contributed. With the planar region knowledge, a set of local transformations can be obtained by constraining matched regions, enabling more precise alignment in the overlapping area. We also use planar knowledge to estimate a transformation field over the whole image. The final mosaic is obtained by a mesh-based optimization framework which maintains high alignment accuracy and relaxes similarity transformation at the same time. Extensive experiments with quantitative comparisons show that our method can deal with different situations and outperforms the state-of-the-arts on challenging scenes.
\end{abstract}

\section{Introduction}

Image stitching is the process of combining multiple photographic images with overlapped field-of-views to produce a wider-view panorama with a higher resolution. It has long been a hot research topic in the community of Computer Vision.
Early methods \cite{ImageAlignmentNStitchingTutorial} are generally built upon global transformation models, e.g. homography, and only work well when the input images follow cylinder projection constraints or the scene can be approximated as a plane with a certain distance to the camera.
When such geometric assumptions are violated, panoramas generated will suffer from misalignments, distortions and ghosting. In order to handle scenes with complex structures or large parallax, spatially varying parametric motion field models are introduced \cite{smoothlyVaryingAffineStitching,APAP} and suited in mesh optimization frameworks for producing desired mosaics \cite{ParallaxTolerantImageStitching,rectanglingPanoramicImagesViaWarping}.
Some works \cite{SPHP,AANAP,Quasi-homographyWarpInImageStitching,NISwGSP} also constrain the model to undergo similarity transformation in order to generate visually pleasing results.

The key challenge of image stitching lies in how to handle images with large parallax. Considering different perspective relationships in such scenes, recent methods have adopted multi-homography to fit the scene \cite{dualHomographyWarp,ImageAlignmentByPiecePlanarRegionMatching,imageStitchUzMultiHomoEstimatBySegRegForDiffParallax,eccv18}. After extracting candidate planes from images, the target image is warped by a global transformation which is the weighted average of multiple local projective transformations, implemented by mesh optimization or other advanced techniques.
However, such methods either are based on image segmentation with heuristic rules or region matching with low-level features or require depth information which needs the special assistance of depth cameras or stereo matching. Under such conditions, the candidate planes they obtain are generally too trivial, unstructured or incomplete, influencing stitching results.

\begin{figure}
	\centering
	\subfigure[\scriptsize Image1]{
		\includegraphics[width=0.22\textwidth]{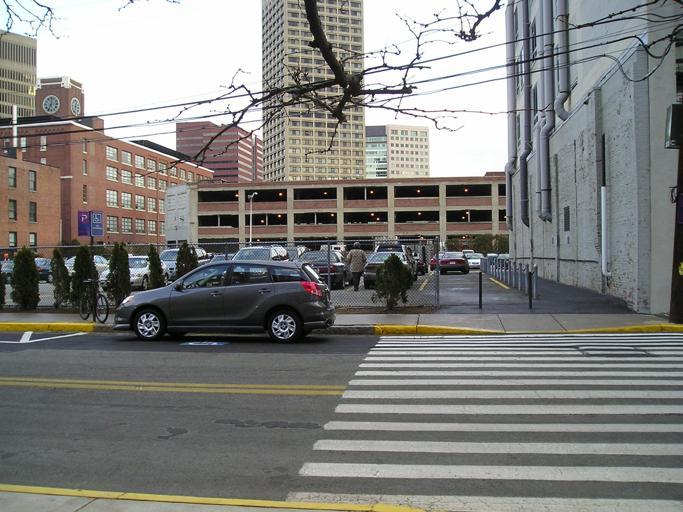}
	}%
	\subfigure[\scriptsize Image2]{
		\includegraphics[width=0.22\textwidth]{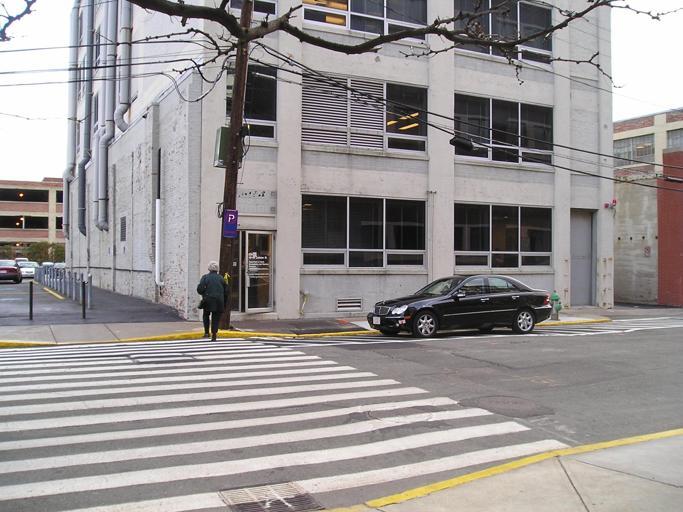}
	}%
	\subfigure[\scriptsize Image1 planar region segmentation]{
		\includegraphics[width=0.22\textwidth]{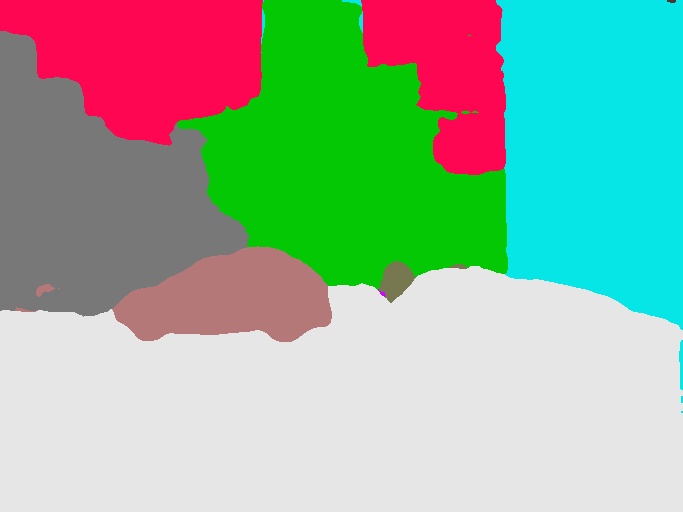}
	}%
	\subfigure[\scriptsize Image2 planar region segmentation]{
		\includegraphics[width=0.22\textwidth]{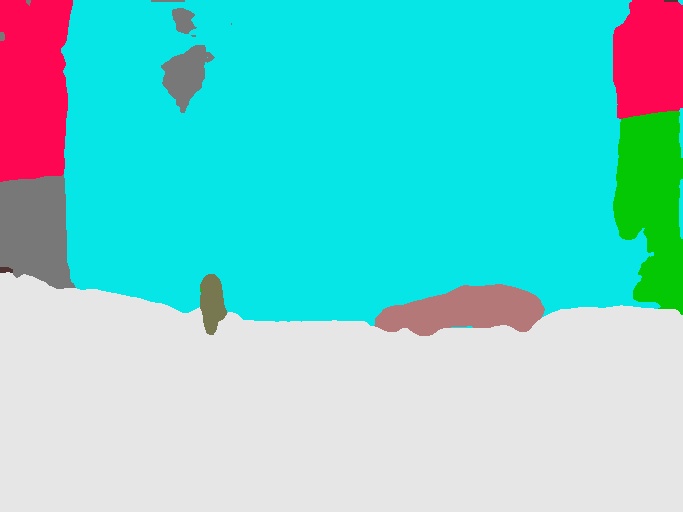}
	}%
	
	\subfigure[\scriptsize ICE]{
		\includegraphics[width=0.23\textwidth]{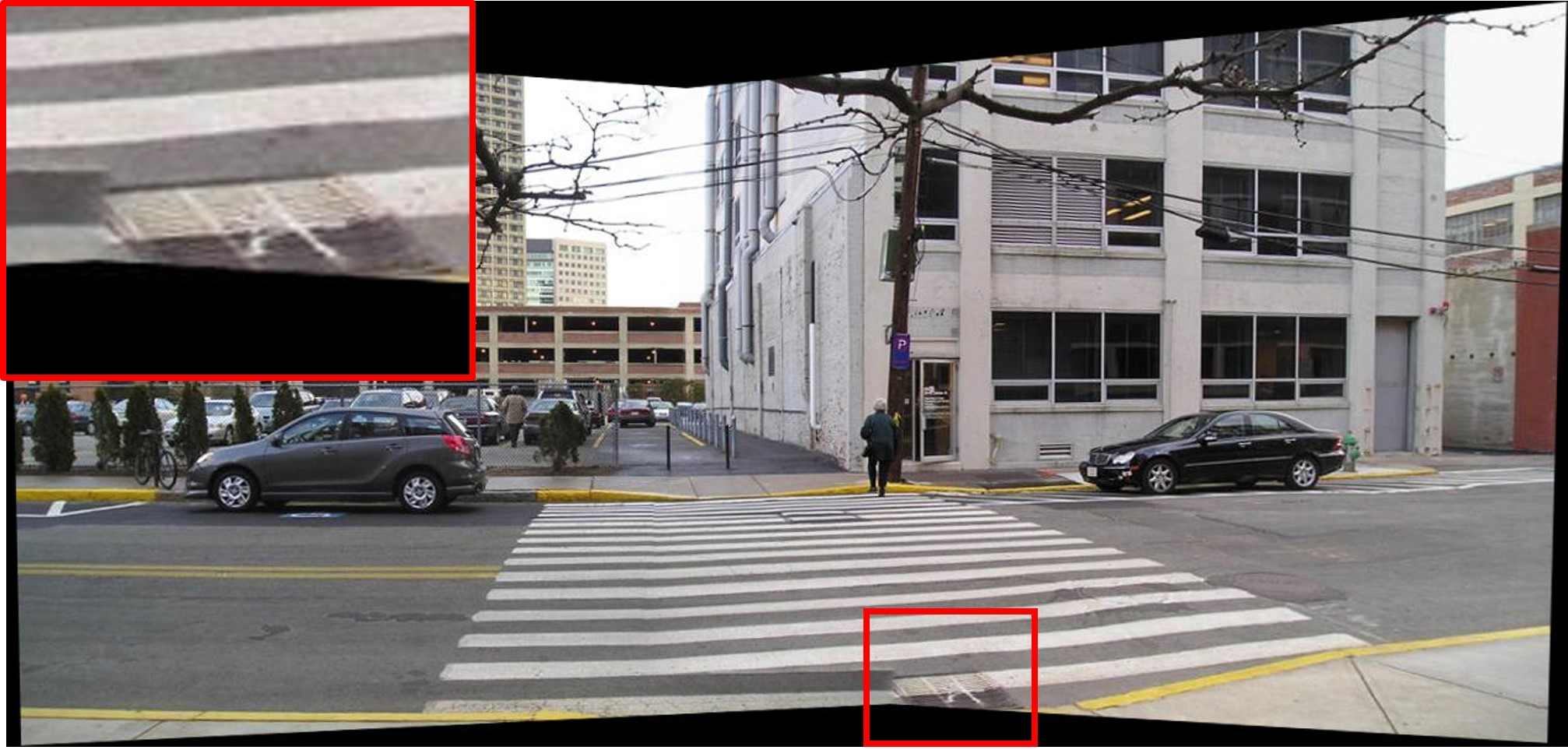}
	}%
	\subfigure[\scriptsize APAP]{
		\includegraphics[width=0.23\textwidth]{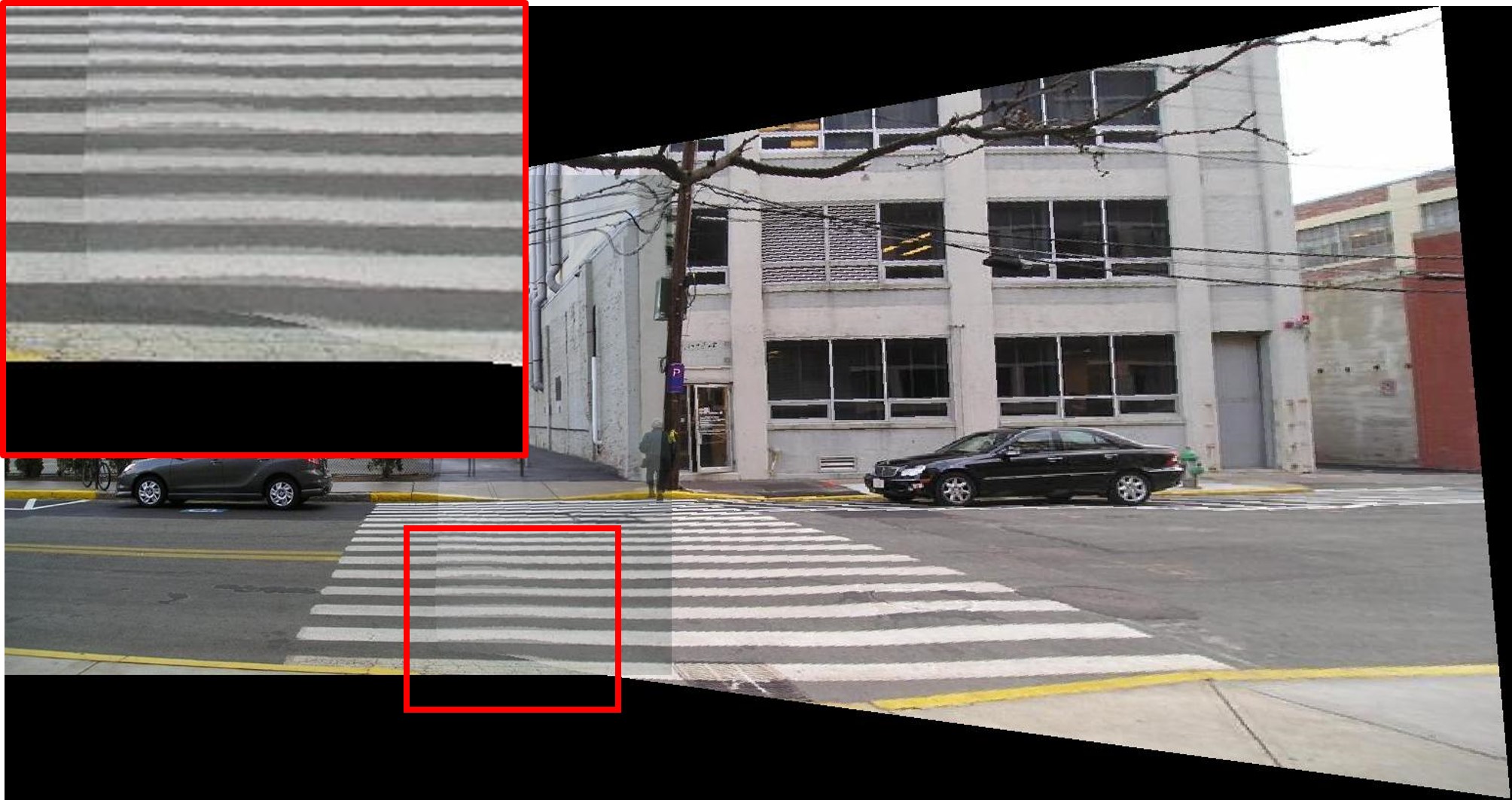}
	}%
	\subfigure[\scriptsize SPHP]{
		\includegraphics[width=0.23\textwidth]{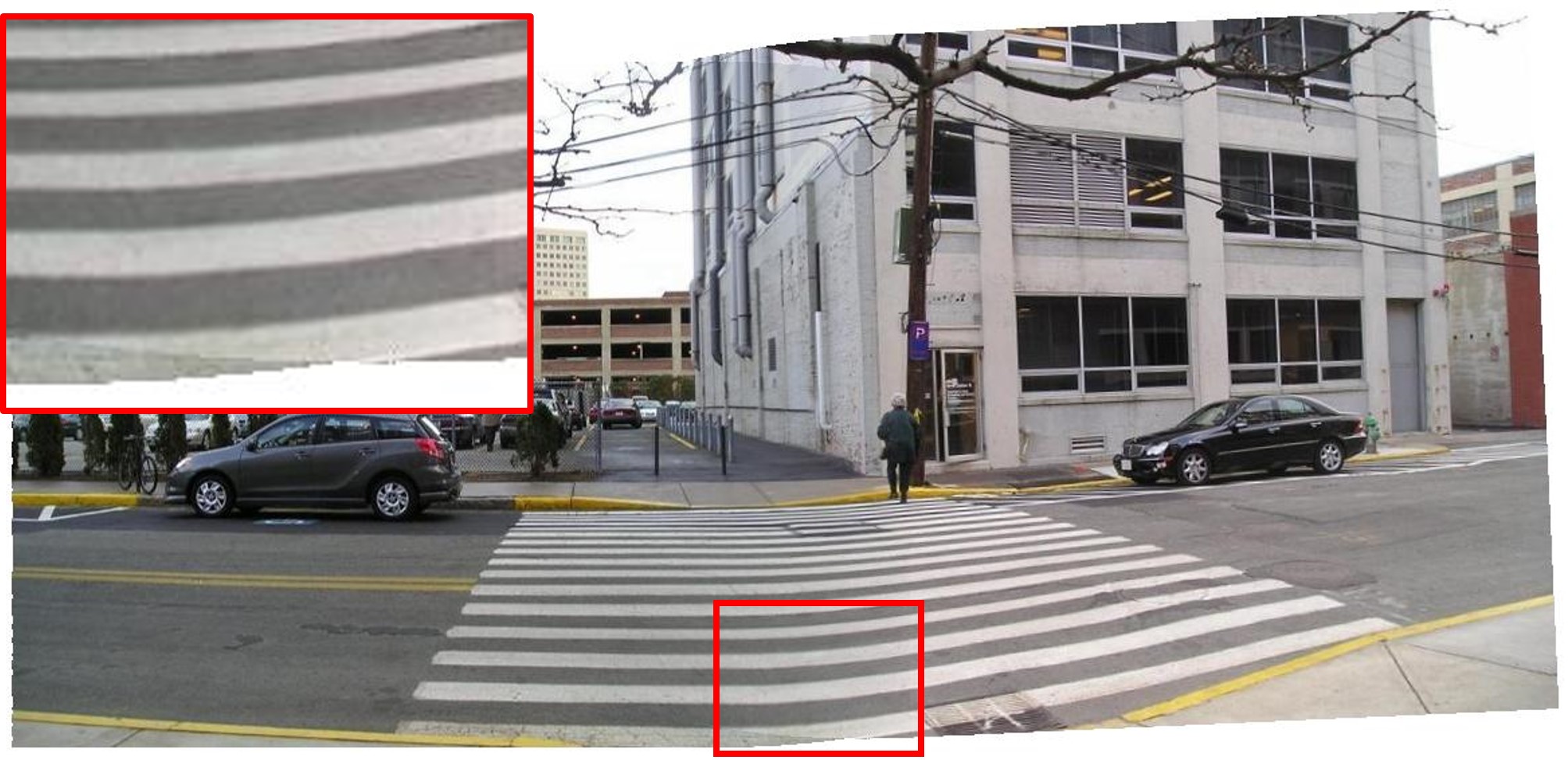}
	}%
	\subfigure[\scriptsize Ours]{
		\includegraphics[width=0.23\textwidth]{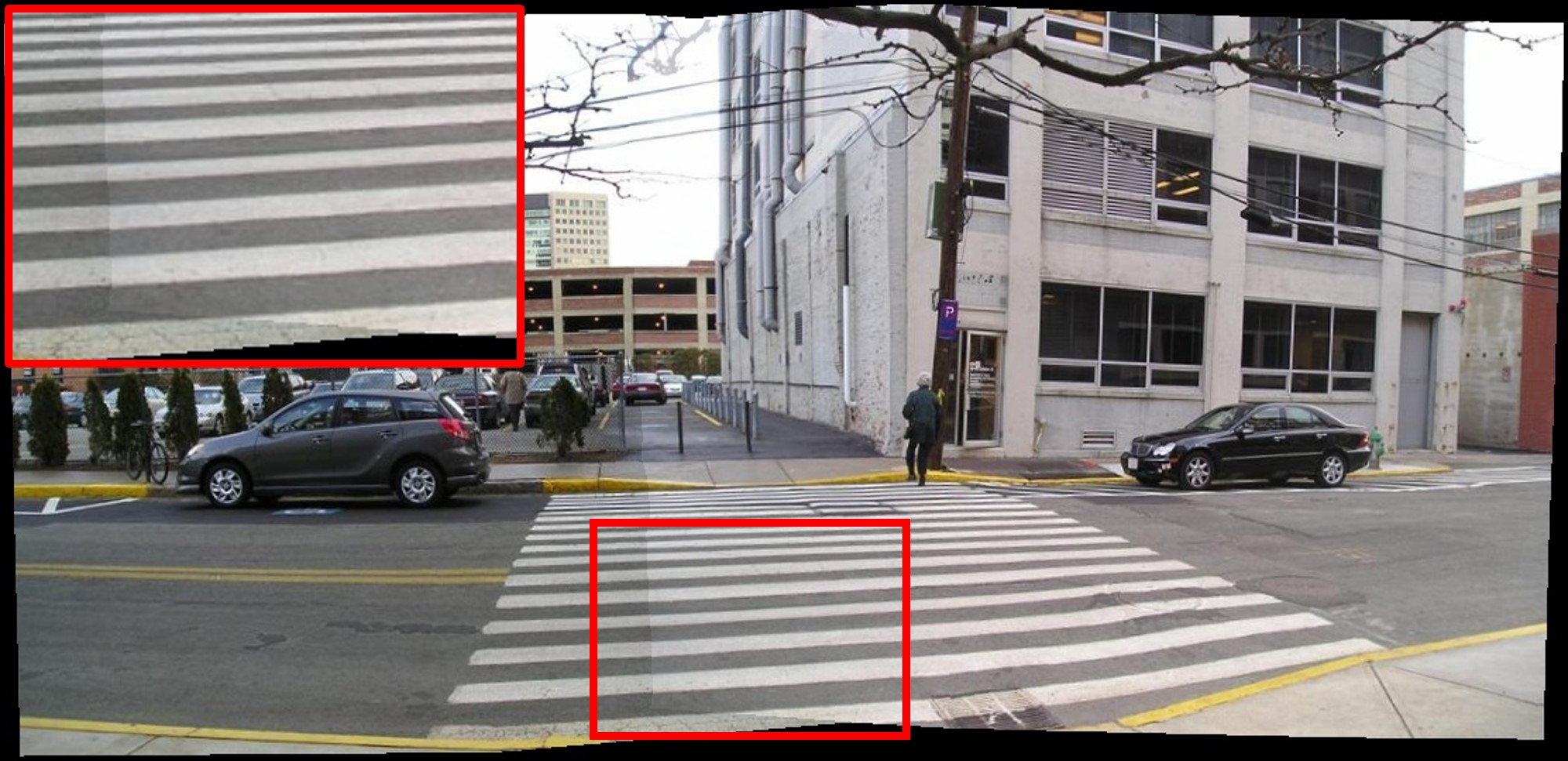}
	}%
	
	\caption{Example of how planar region segmentation helps stitching. In the input image pair (a) and (b), there are three matched dominant regions, i.e., the ground plane, the near white building, and the far one whose positions vary greatly from each other. With feature matching on these planar regions between the image pair, our method finds more accurate local transformation models and achieves better alignment. By contrast, other methods based on global matching have difficulty in finding all valid features, leading to misalignments or bending of straight lines, as highlighted in (e)-(g).}
	\label{planarSegDemo}
\end{figure}

In this paper, we propose the concept of \textbf{planar region} based on which a new image stitching method considering planar region consensus is developed. Different from the traditional plane, a planar region on an image is a set of pixels whose real world correspondences roughly reside on the same plane. Fig. \ref{planarSegDemo} (c) and (d) show two planar segmentation results on an image pair.
Since a planar region should be consistent under different views, the number of matched features in consistent planar regions should be significantly more than
the number of matched features
in non-consistent ones. Therefore, planar region consensus can work as a good constraint which would facilitate determining valid local models as well as better global transformation.

We argue that rich semantic information directly extracted from RGB images can be used for planar region detection by advanced semantic segmentation architectures. However, since this task essentially can be viewed as a clustering problem whose region labels only have latent semantics, directly applying existing networks would not suffice. Instead of contriving new architectures, we base our model on the well-known scene parsing network UPerNet \cite{UperNet101} and add an extra layer to its output end for computing the clustering loss. For each sample, we calculate a new permuted mask of ground truth for back propagation through maximum weighted matching which preserves clustering properties, while giving a reasonable clustering loss. In this way, we improve the performance greatly. Besides, since we only add an extra layer to existent networks, our scheme can be incorporated with different architectures to build various end-to-end models.

With planar regions detected, we stitch images under their guidance by finding matched region pairs and further calculating matched vertices and regional transformations based on local transformation models. The mosaic is then generated via a mesh optimization framework with an objective function specifically designed to fully respect local models, while maintaining smoothness and continuity of global transformation at the same time. We show the robustness of our method and its superiority over the state-of-the-arts in handling complex scene geometry and camera motions with extensive experiments.

Our paper makes the following contributions.
\begin{itemize}
	\setlength{\leftmargin}{2em} %左边界
	\setlength{\parsep}{0ex} %段落间距
	\setlength{\topsep}{0ex} %列表到上下文的垂直距离
	\setlength{\itemsep}{0ex} %条目间距
	\setlength{\labelsep}{1em} %标号和列表项之间的距离,默认0.5em
	\setlength{\itemindent}{0em} %标签缩进量
	\setlength{\listparindent}{0em} %段落缩进量
	\item[(1)] We propose the concept of planar region, build a dataset for planar region segmentation, and for the first time use a deep Convolutional Neural Network to extract planar regions from RGB images directly.
	\item[(2)] We propose to stitch images with a new mesh optimization framework which respects the constrains coming from planar region consensus, allowing for local alignments of matched planes and meanwhile ensuring global smoothness with new energy terms introduced.
\end{itemize}
\section{Related work}

Szeliski et al. \cite{ImageAlignmentNStitchingTutorial} gave a comprehensive survey on traditional image stitching. However, since global alignment models cannot account for images with large parallax, spatially-varying methods have gained growing popularity. Lin et al. \cite{smoothlyVaryingAffineStitching} developed a spatially varying affine stitching field to align images. Zaragoza et al. \cite{APAP} proposed an as-projective-as-possible warp which interpolates a smoothly varying projective mesh to guide stitching. Li et al. \cite{DualFeatureWarp} proposed to align images using line features as well as feature points and obtained better results in low-texture areas.
Chen et al. \cite{NISwGSP} introduced a global similarity prior into optimization framework. Through proper selection, this term lessens accumulate error from stitching multiple images and helps generate more natural panorama. Lin et al. \cite{directPhotoAlignByMeshDefo} incorporated mesh-based image warping with dense photometric alignment on sampled pixels for image stitching and video stabilization.
By providing functional and effective terms, warped mesh can preserve salient content in the image and improve stitching quality significantly.

Another branch of image stitching aligns the input images through a series of seams whose positions are determined by energy minimization on pixel level. Gao et al. \cite{seamDrivenImageStitching} used seams to find a better motion model for stitching. Lin et al. \cite{SEAGULL} explicitly used seams to guide local alignment optimization process to iteratively improve the stitching result. He et al. \cite{parallaxRobustSurveillanceVideoStitching} extended alignment to video stabilization incorporating layered warping and the change-detection based seam updating. Though seam is able to handle scenes of big parallax,
the cost of calculating optical flow is always high compared with mesh-based methods.

Segmenting images into several regions of interest has also been investigated a lot. Felzenszwalb et al. \cite{efficientGraphBasedImageSegmentation} used graph to segment images into superpixels with similar color intensity in linear time.
In recent years, segmentation with semantic or planar meanings gained more attention. Shelhamer et al. \cite{fullyConvNetworksForSemanticSeg} firstly used convolution networks to tackle semantic segmentation and achieved $20\%$ relative improvement at the time. Ronneberger et al. \cite{UNET} invented a U-shaped net which concatenates information across different layers to preserve more information for small datasets. Xiao et al. \cite{UperNet101} introduced the concept of unified perceptual parsing and built a network with hierarchical structure to segment at multiple perceptual levels. These works have increased the performance of semantic segmentation and scene understanding greatly. For plane segmentation, Lin et al. \cite{dualHomographyWarp} segmented an image roughly into the distant plane and the ground plane through clustering on keypoints and weighted averaging. Zheng et al. \cite{PCPS} used detected keypoints as vertexes of a triangulated mesh from which a plane segmentation is formulated. Liu et al. \cite{PlaneNet} designed PlaneNet to solve plane detection and parameter and depth estimation from a single RGB image under a united framework. Liu et al. incorporated object detection networks to enable segmenting any number of planes from the image \cite{PlaneRCNN}. However, the planes these neural networks detect are rigidly planar which would incur over-segmentation and obstruct the discovery of useful matching patterns for image stitching.

\section{Planar Region Segmentation}

People can easily recognize planar regions directly from RGB images with semantic information and global context. For example, the outer wall of a building can be roughly approximated as a plane, while different sides compose different planar regions. In this sense, planar regions represent a certain level of semantics and could be certainly detected with the aid of mature semantic segmentation networks.
Since the planar regions have only latent semantics and label information with specific semantic classes is not required for image stitching,
what we really need is a clustering result on image pixels under the coplanar constraint.
By comparison, deep semantic segmentation networks generally require clear classification labels to work.

\begin{figure}
	\centering
	\includegraphics[width=1.0\textwidth]{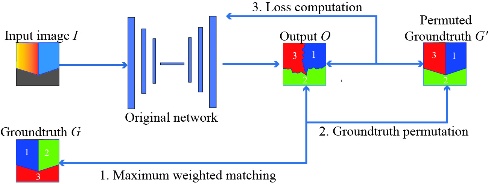}
	\caption{Illustration of our network architecture and its behavior in training input image $I$. Our scheme can be integrated with any network architectures by adding an extra block which computes the permuted ground truth mask $G^\prime$ for loss calculation and back propagation.}
	\label{networkScheme}
\end{figure}

The main issue for clustering with networks on semantic segmentation is that the criteria for clustering is intrinsically different from classification. To handle this, we need to design a new module which is derivable and allows for calculating an appropriate loss of clustering. Instead of designing a new architecture completely, we basically build our end-to-end model on UPerNet\cite{UperNet101} due to its superior performance on scene understanding, only modifying the back propagation module to accommodate planar region segmentation.

Our motivation comes from the property of clustering. As shown in Fig. \ref{networkScheme}, permuting the labels of ground truth yields an equivalent ground truth under clustering, without the need of paying attention to specific class labels. For this reason, we can always permute the labels of ground truth $G$ according to the semantic network output $O$ to obtain a desired segmentation mask without introducing error. Ideally, the permuted ground truth $G^\prime$ which has the most same labels with $O$ should be the closest to network output and sufficient for back propagation directly through classification loss. To obtain the permutation, we first construct the bipartite graph $B$ whose weight matrix is the confusion matrix of $O$ and $G$. The permutation is just the maximum weighted matching of $B$.
Based on the permuted groundtruth $G^\prime$ and the original output $O$, we calculate their similarity and use cross entropy as loss for back propagation.

It should be noted that the method is derivable since the matching and permutation can be accomplished outside the network. Using the idea described above, we add an extra block calculating the permuted ground truth for back propagation. The model with our extra module appended still works in an end-to-end manner.

% \IncMargin{1em} % 使得行号不向外突出
\begin{algorithm}
	\SetAlgoNoLine % 不要算法中的竖线
	\SetKwInOut{Input}{\textbf{Input}}\SetKwInOut{Output}{\textbf{Output}} % 替换关键词
	
	\Input{
Output segmentation mask $O$, ground truth mask $G$
	}
	\Output{
		Permuted ground truth mask $G^\prime$}
	\BlankLine
	
	Calculate confusion matrix $C$ from $O$ and $G$ with $C_{ij}=|\{(m,n)|O(m,n)=i\wedge G(m,n)=j\}|$;
	
	Build a bipartite graph $B=(V,E)$ where $V=\{v_i|i\in O\vee G\}$ and $E=\{(v_i,v_j,C_{ij})|i\in O\wedge j\in G\}$;
	
	Find maximum weighted matching $M$ on $B$;
	
	Construct permuted ground truth segmentation mask $G^\prime$ following
	$G^\prime(m,n)=M(G(m,n))$.
	\caption{Find the permuted mask for loss computation \label{planarMaskLayerAlgo}}
\end{algorithm}
\DecMargin{1em}

\begin{figure}
	\centering
	\includegraphics[width=1.0\textwidth]{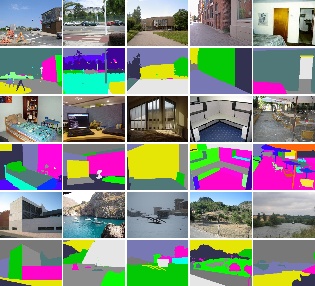}
	\caption{Images and segmentation masks from our planar region segmentation dataset.}
	\label{segDataset}
\end{figure}

Since an image usually contains a limited number of planar regions, the maximum weighted matching runs very fast, and the training process can proceed without incurring too much computation overhead. Besides, our scheme operates independently of the network and can be integrated with any architecture representing the state-of-the-arts. \\
\textbf{Dataset.} Since the concept of planar region is newly proposed and no datasets are available so far, we manually relabel 1005 images from ADE20k scene parsing dataset \cite{zhou2017scene} upon the semantic segmentation groundtruth by splitting multi-planar blocks and merging coplanar objects (as shown in Fig. \ref{segDataset}).  We use 795 images (approximately $79\%$) for training and the rest for validation. Our dataset will be released for future research.

With the scheme proposed above, we achieve a higher accuracy and mean intersection-over-union (IoU) than the original UPerNet on this task. Interestingly, results of our planar segmentation net still exhibit certain semantics. We observe that the sky and ground plane are always assigned the same label in different images, and buildings always share similar labels even if the numbers of region vary significantly in a few cases. This observation suggests that latent semantics are also learned by our network. Results are discussed in Section 5.2.

\section{Image Stitching with Planar Region Consensus}
To make fully use of matched planar region knowledge, we introduce a new image stitching method. The key is to make the stitching process respect the constraint of planar region consensus. We rely on this to harvest more accurate matched regions so as to estimate more accurate local transformations and smooth global transformation. Image stitching is accomplished through a newly developed mesh optimization framework framed as regional information and newly-designed optimization terms. The key stages include: finding matched regions, generating matched points, estimating the transformation field, mesh optimization, and image stitching by texture mapping. 

Here the segmentation mask has been calculated. For each one of $n$ input images to be stitched,  we construct a regular quad mesh. The whole mesh $V$ is constituted by $\{(V_i, E_i)|i=1,...,n\}$, with $V_i$ and $E_i$ denoting separately the vertex set and edge set of the image $I_i$.

\subsection{Finding matched regions}

\begin{figure}
	\centering
	\includegraphics[width=1.0\textwidth]{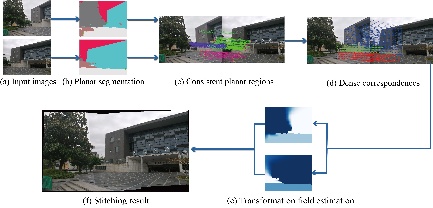}
	\caption{Our image stitching pipeline based on planar region consensus. We first apply planar region segmentation to the input image pair. Regional RANSAC can then be applied, yielding a set of consistent planar regions based on which we obtain the dense correspondences and estimate the transformation field. The final mosaic is obtained by a mesh optimization framework that fully utilizes regional information.  }
	\label{PRERASCDemo}	
\end{figure}

For images with overlapped content, we believe the number of matched keypoints in regions coming from the same object in the real world exceeds
the number of those not belonging to the same one greatly. We use this as a constraint to find valid consistent regions. Since outliers still
exist, we apply RANSAC \cite{RANSACOrig} to find reliable matching in all candidate region pairs. Intuitively, the bigger the number of matched keypoints is,
the more likely two regions are from the same object. We can thus safely use the matching number of keypoints as a weight for each candidate region pair.
To find all matches, we apply maximum weighted matching on all region pairs which yields a group of most probably consistent planar regions.
We name the above method as regional RANSAC.

Fig. \ref{PRERASCDemo} showing our image stitching pipeline gives an example where three groups of matched regions are extracted using Regional RANSAC. Since the number of matched planar regions in an image pair generally does not exceed 10, the time used for maximum weighted matching can be ignored. In this way, the total running time of computing regional RANSACs will not exceed the running time of global RANSAC.

\subsection{Gathering regional information}
Mostly, the regions which are found consistent by the above process dominate the overlapping area of images to be stitched. We thus call them dominant regions which play crucial roles in stitching. Intuitively, matched dominant regions give strong alignment cues. The above regional RANSACs, however, only bring us a set of reliable but sparse feature correspondences. Induced by them, we aim to densify the correspondences by figuring out the matched positions for all mesh vertices of dominant regions in the overlapping area, which would facilitate mesh optimization for obtaining more precise stitching. \\
\textbf{Dense correspondence establishment.} Based on the sparse feature correspondences yielded by regional RANSACs above, we compute its counterpart for each mesh vertex by a scheme similar to APAP\cite{APAP}. We use $M_{ij}$ to denote the set of correspondences thus obtained which offers alignment information for mesh vertices. An example is given in Fig.\ref{PRERASCDemo}(d), where three groups of matched point pairs are obtained. The dense correspondences obtained should be respected during mesh optimization. We call it dense correspondence constraint.

Apart from alignment, matched dominant regions also encode certain semantic information. By keeping transformation consistency between each pair of matched regions, salient structures shown in the dominant planar regions can be preserved through encouraging the region to undergo the same transformation. We offer this regional information by estimating a transformation field covering all mesh vertices. Specifically, we estimate the similarity transformation since it incurs little distortion.

\textbf{Transformation field estimation.}
For a dominant region $i$, the similarity transformation it undergoes can be expressed as
\begin{equation}
S_i(s,\theta)=\left(
\begin{array}{cc}
s\cos\theta & s\sin\theta \\
-s\sin\theta & s\cos\theta \\
\end{array}
\right)
\end{equation}
where $s$ and $\theta$ are the scale and rotation angle, separately. We use $c_i=s\cos\theta$ and $s_i=s\sin\theta$ in the following for a linear expression. This transformation can be estimated directly using its regional matched SIFT keypoints.

For a vertex which lies in a dominant region, we assign its transformation to be the same as the region. For other vertices without planar constraint, we can easily generate their transformations by using the inverse distance weighted interpolation. Fig. \ref{PRERASCDemo}(e) shows our estimation result. It is obvious that a smooth transition field is established.

\subsection{Mesh optimization}
With the regional information yielded by planar region consensus, we are ready to construct our mesh optimization framework which produces natural stitching with accurate alignment in the overlapping area. 
Our objective function for mesh optimization consists of four terms. They are 
the alignment term $E_a$, regional similarity term $E_r$, local similarity term $E_s$, and line preserving term $E_l$. \\
\textbf{Alignment term $E_a$.} This term ensures alignment quality by encouraging the mesh to obey dense correspondence constraint and is calculated as
\begin{equation}
E_a(V)=\sum_{i,j}\sum_{\{v_m,v_n\}\in M_{ij}}\left\|\phi(v_m)-\phi(v_n)\right\|^2
\end{equation}
where $v_m$ and $v_n$ are two points with correspondence in $I_i$ and $I_j$.  $\phi(v)=\sum_{i=1}^{4}\alpha_iv_i$ expresses $v$ as the linear combination of the four corner vertices of the quad surrounding it, with $\alpha_i$ being the bilinear weight. \\
\textbf{Regional similarity term $E_r$.} For each individual vertex, we want its transformation to follow the estimated transformation field in order to preserve properties of planar regions. With this term added, trivial solution can be avoided effectively and regional consistency is fully respected. The term is defined as
\begin{equation}
E_r(V)=\sum_{i}\sum_{(v_j,v_k)\in E_i}[(c(e_{jk})-c(v_j))^2+
(s(e_{jk})-s(v_j))^2
]
\end{equation}
where $c(v_j)$ and $s(v_j)$ represent the parameters of similarity transformation pre-computed in the above subsection. $c(e_{jk})$ and $s(e_{jk})$ denote parameters of the similarity transformation edge $e_{jk}$ undergoes. Let $S_{jk}$ be the transformation. It is parameterized as
\begin{equation}
S_{jk}=\left(
\begin{array}{cc}
c(e_{jk}) & s(e_{jk}) \\
-s(e_{jk}) & c(e_{jk}) \\
\end{array}
\right)
\end{equation}
where the coefficients are expressed as linear combinations of vertices \cite{implementingAsRigidAsPossible}.
\\
\textbf{Local similarity term $E_l$.} To enforce the mesh to undergo similarity transformation in each quad, we adopt the same constraint as \cite{NISwGSP}
\begin{equation}
E_l(V)=\sum_{i}\sum_{(v_j,v_k)\in E_i}\left\|
(\tilde{v_j}-\tilde{v_k})-S_{jk}(v_k-v_j)
\right\|^2
\end{equation}
where $v_j$ and $v_k$ represent the vertices of edge $e_{jk}$ in $I_i$. $\tilde{v_j}$ and $\tilde{v_k}$ denote the transformed positions of $v_j$ and $v_k$.  \\
\textbf{Line preserving term $E_l$.}  Human are more sensitive to salient linear structures in the scene, for which relevant terms have been investigated \cite{directPhotoAlignByMeshDefo,SEAGULL}. Here we adopt the term to preserve straight lines. After detecting line segments $L$ using the detector \cite{LSD:ALineSegmentDetector}, we uniformly sample keypoints along the line. Each keypoint has a 1D local coordinate $a\in[0,1]$ relative to the two endpoints of the line segment it resides. This coordinate should hold the same after deformation. We define the constraint as
\begin{equation}
E_l(V)=\sum_{i}\sum_{l_{uv}\in L_{i}}\sum_{l_k\in l_{uv}}
\left\|\phi(l_k)-((1-a)\phi(l_u)+a\phi(l_v))\right\|^2
\end{equation}
where $l_u$ and $l_v$ are the endpoints of the segment with $l_k$ being the sampled keypoint.

The optimal mesh $\widetilde{V}$ is determined by
\begin{equation}
\widetilde{V}=\arg\min_{V} \lambda_aE_a(V)+E_r(V)+\lambda_s E_s(V)+\lambda_lE_l(v)
\label{naturalEnergy}
\end{equation}
where $\lambda_a$, $\lambda_s$ and $\lambda_l$ control the importance of different terms and are set to $0.12$, $0.08$, and $0.3$ separately in our experiments. The optimization can be efficiently solved by a sparse linear solver. With the mesh calculated, the final image is synthesized by deforming the input images under the guidance of solved mesh using texture mapping. In our experiments, the total stitching process can be finished within one minute.

\section{Experiments}
\begin{figure}
	\centering
	\includegraphics[width=1\linewidth]{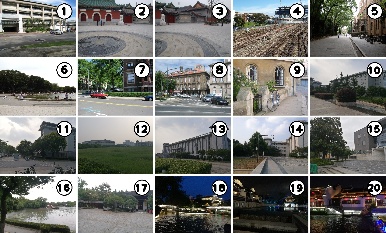}
	\caption{Images used for quantitative evaluation. Image pairs 1-6 are from \cite{dualHomographyWarp,APAP,SPHP}, 7-9 are from \cite{zhou2017scene}, and 10-20 are taken by ourselves.}
	\label{dataset}
\end{figure}
We name our image stitching method Planar Region Consensus Stitching, abbreviated as $PRCS$. We compare with ICE\footnote{https://www.microsoft.com/en-us/research/product/computational-photography-applications/image-composite-editor/}, APAP\cite{APAP}, SPHP\cite{SPHP}, NISwGSP\cite{NISwGSP} and RISwMR \cite{eccv18} which represent the state-of-the-arts. For evaluation, we compare our methods on a combined image dataset of 20 images as shown in Fig. \ref{dataset}. It includes image pairs from works before, ADE20k and those captured by ourselves. Since some methods do not release intermediate results, quantitative evaluation is only done with APAP, SPHP and NISwGSP. Please see our accompanying material for all the experimental results.

\subsection{Experiment settings}
All code is written in Python. For planar region segmentation, we adopt UPerNet as the inner architecture, and the code is fetched directly from \cite{zhou2018semanticCode} except that we redesign and reimplement the loss back propagation layer with Algorithm \ref{planarMaskLayerAlgo}.
The keypoints are extracted with SIFT \cite{SIFT} from OpenCV. The meshes used for optimization for all our testing images are $100\times 100$. The mesh is solved with the sparse linear solver from SciPy.
\subsection{Quantitative Evaluation}
\begin{table}
	\centering
	\caption{Accuracy and mean IoU (intersection-over-union) in percentage of planar segmentation on different selections of $K$, the maximum number of planes and implemented w/o our specifically designed back propagation module.}
	\begin{tabular}{|c|c|c|c|c|}
		\hline
		$K$ & acc(w/o) & acc(w) & IoU (w/o) & IoU (w) \\
		\hline
		12 & 57.65 & \textbf{82.63} & 38.85 & \textbf{69.95}  \\
		16 & 57.11 & \textbf{81.95} & 38.80 & \textbf{71.27}  \\
		20 & 56.02 & \textbf{81.42} & 35.71 & \textbf{67.84} \\
		\hline
	\end{tabular}
	\label{planarSegResults}
\end{table}

For planar segmentation, we compare the accuracy and mean IoU on the segmentation results with/without our newly designed layer for loss back propagation. As shown in Table \ref{planarSegResults}, our algorithm greatly improves both scores on this task. Since different selections of $K$ do not have great influence on segmentation quality, we choose $K$ = 16 which is sufficient for most scenes.

For image stitching, since ground truth is not available for non-overlapping regions, we evaluate the stitching quality by measuring the similarity in the overlapping area after deformation. Specifically, we adopt the measure proposed in \cite{DualFeatureWarp,directPhotoAlignByMeshDefo}. We compute the RMSE of one minus normalized cross correlation (NCC) over a neighborhood $\pi$ of $5\times 5$ window in the overlapping area. In this way, the score for the generated mosaic is defined as
\begin{equation}
Score(I_{tar},I_{ref})=\sqrt{\frac{1}{N}\sum_{\pi}(1.0-NCC(p_{ref},p_{tar}))^2}
\end{equation}
where $N$ is the pixel number in the overlapping area. $p_{ref}$ and $p_{tar}$ denote the corresponding pixels of the two input images, respectively.

\begin{figure}
	\centering
	\subfigure[\scriptsize Global RANSAC]{
		\begin{minipage}{0.32\textwidth}
			\centering
			\includegraphics[width=1.0\textwidth]{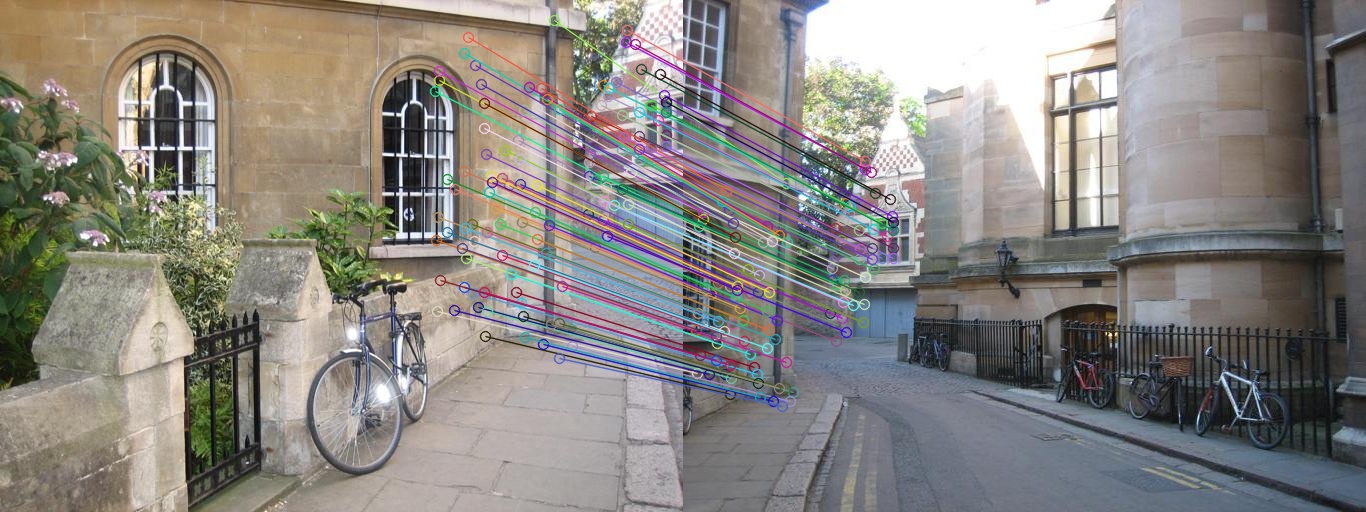}
			%\caption{fig2}
		\end{minipage}
	}%
	\subfigure[\scriptsize Regional RANSAC]{
		\begin{minipage}{0.32\textwidth}
			\centering
			\includegraphics[width=1.0\textwidth]{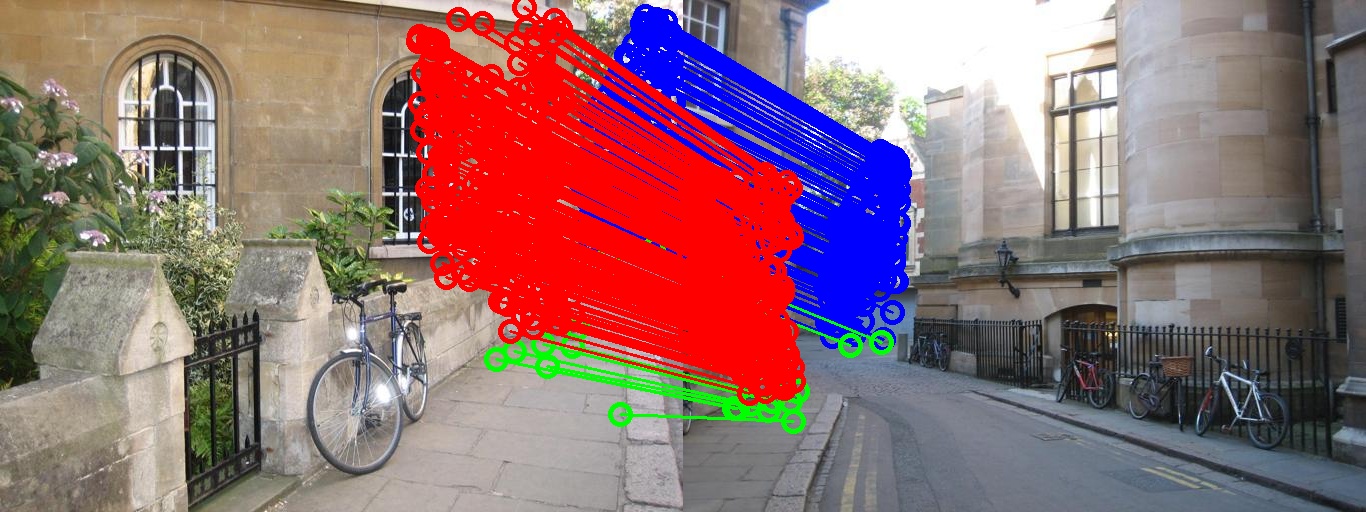}
			%\caption{fig2}
		\end{minipage}
	}%
	\subfigure[\scriptsize APAP]{
		\begin{minipage}{0.32\textwidth}
			\centering
			\includegraphics[width=0.90in]{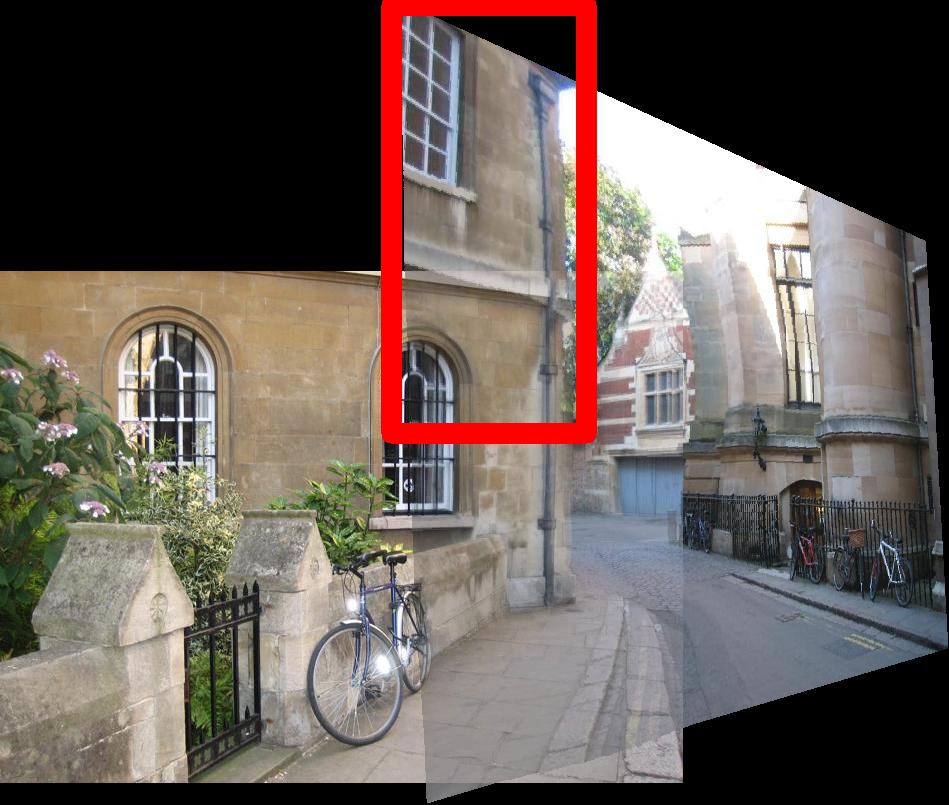}
			%\caption{fig2}
		\end{minipage}
	}%
	
	\subfigure[\scriptsize ICE]{
		\begin{minipage}{0.32\textwidth}
			\centering
			\includegraphics[width=1.21in]{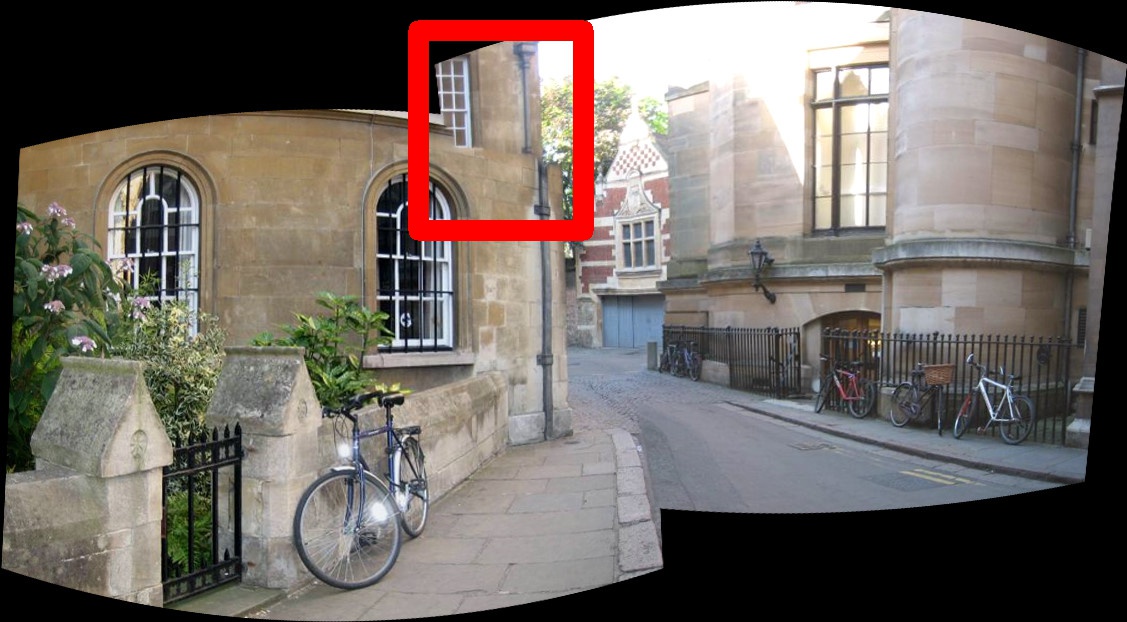}
		\end{minipage}
	}%
	\subfigure[\scriptsize NISwGSP]{
		\begin{minipage}{0.32\textwidth}
			\centering
			\includegraphics[width=1.16in]{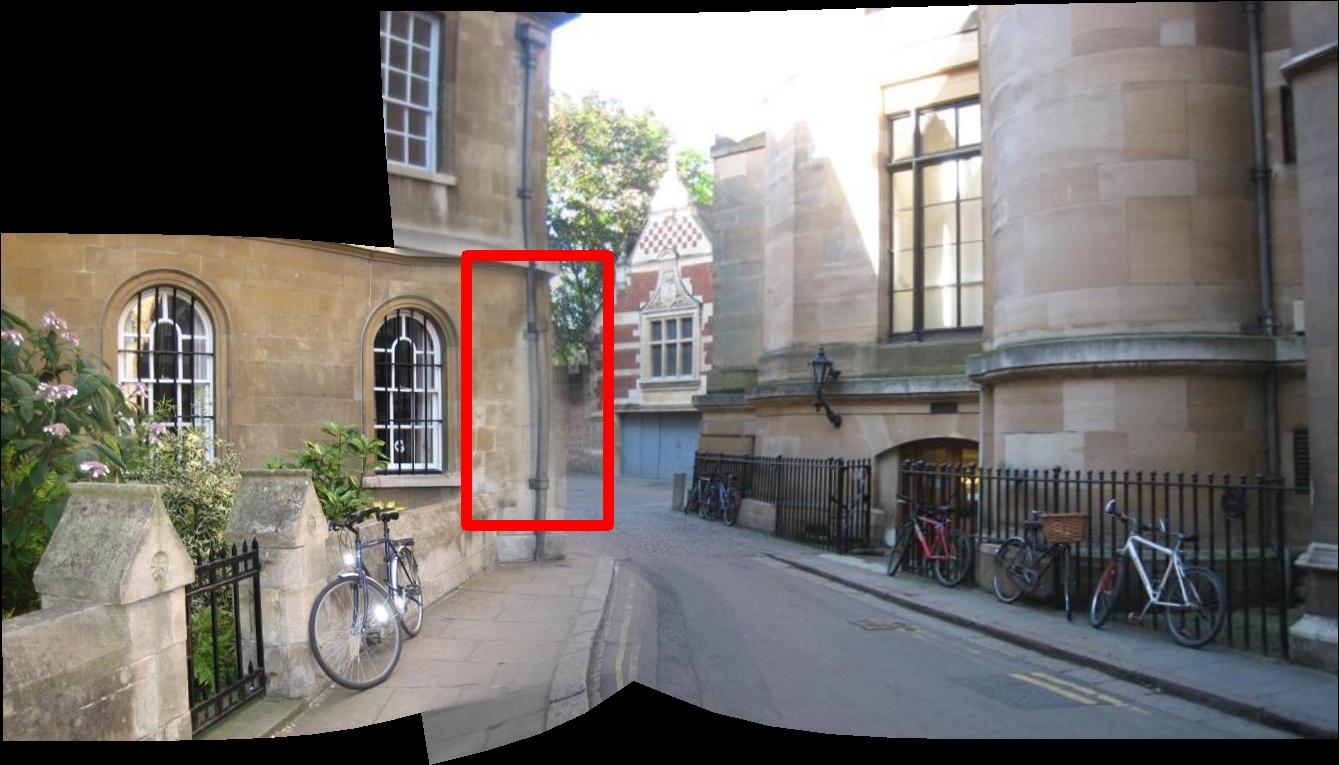}
		\end{minipage}
	}%
	\subfigure[\scriptsize Ours]{
		\begin{minipage}{0.32\textwidth}
			\centering
			\includegraphics[width=1.25in]{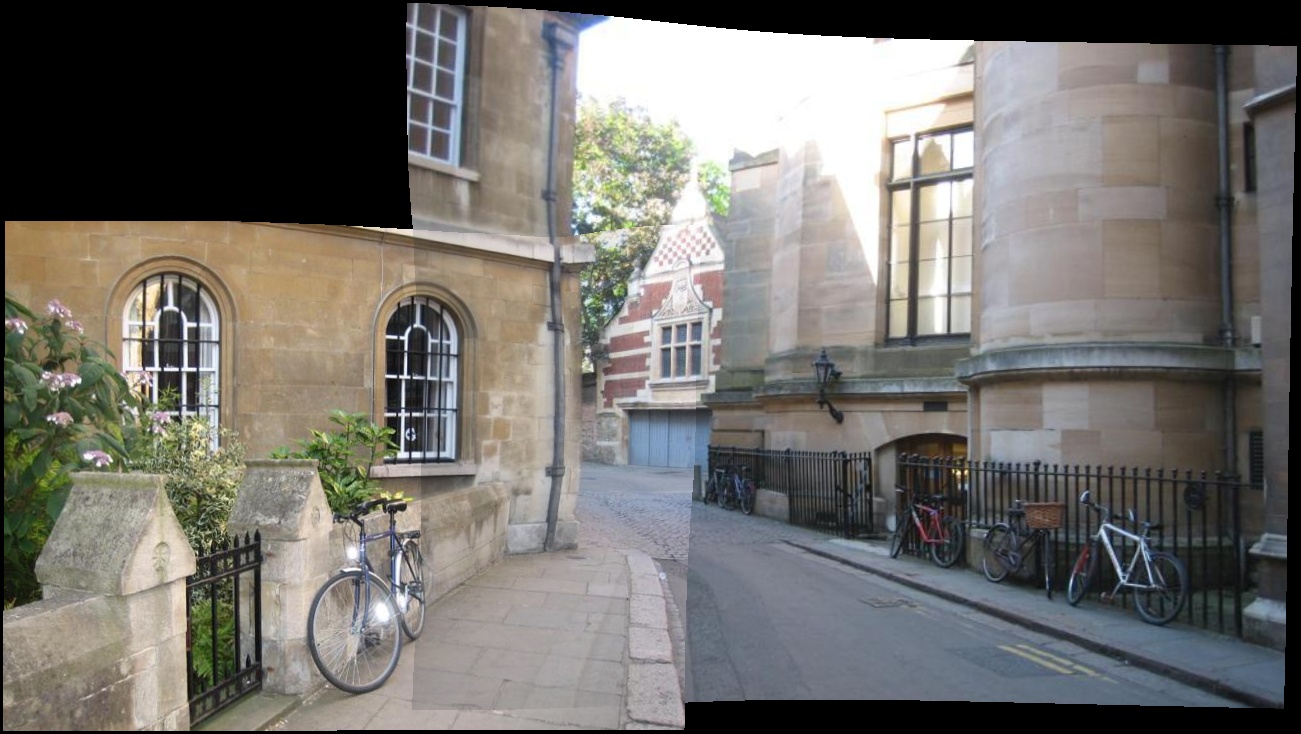}
		\end{minipage}
	}%
	
	\caption{Corner image pair (from \cite{zhou2017scene}).  Artifacts are highlighted. We extract three matched dominant planar regions and deal with them respectively. Only our method maintains the consistency of the left yellow wall.}
	\label{cornerAnalysis}
\end{figure}

\begin{figure}
	\centering
	\subfigure[\scriptsize Global RANSAC]{
		\begin{minipage}{0.32\textwidth}
			\centering
			\includegraphics[width=1.0\textwidth]{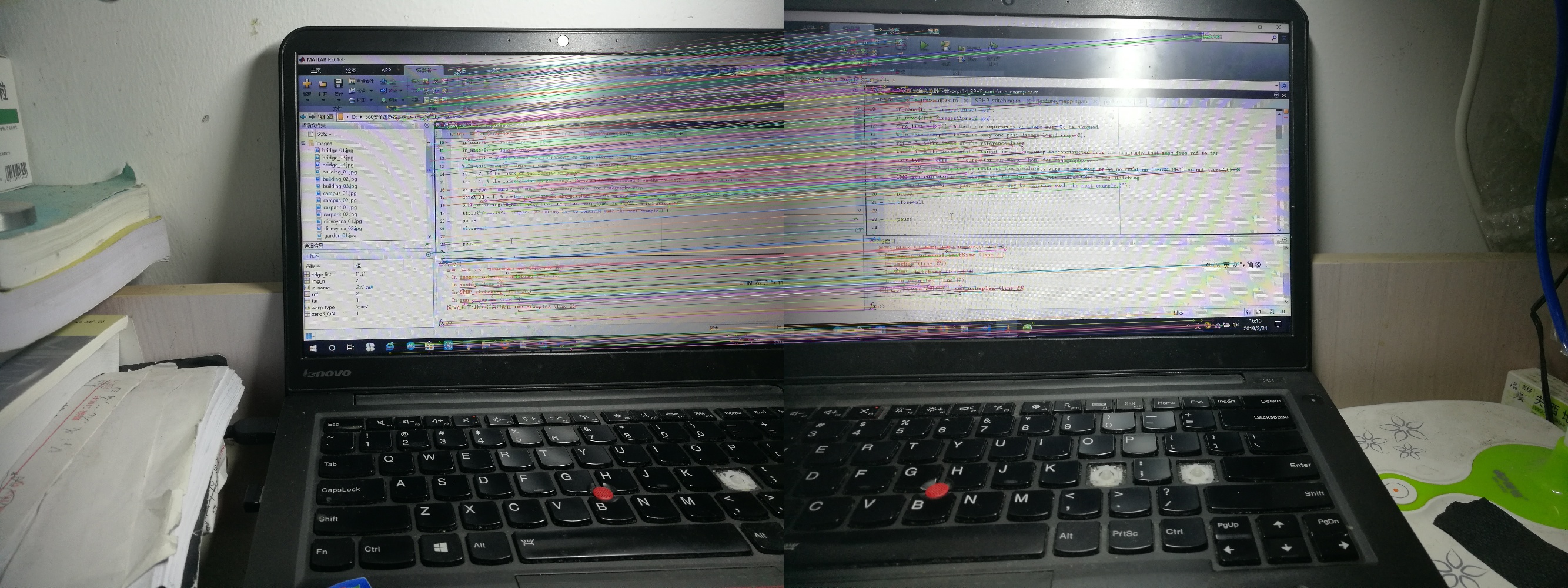}
			%\caption{fig2}
		\end{minipage}
	}%
	\subfigure[\scriptsize Regional RANSAC]{
		\begin{minipage}{0.32\textwidth}
			\centering
			\includegraphics[width=1.0\textwidth]{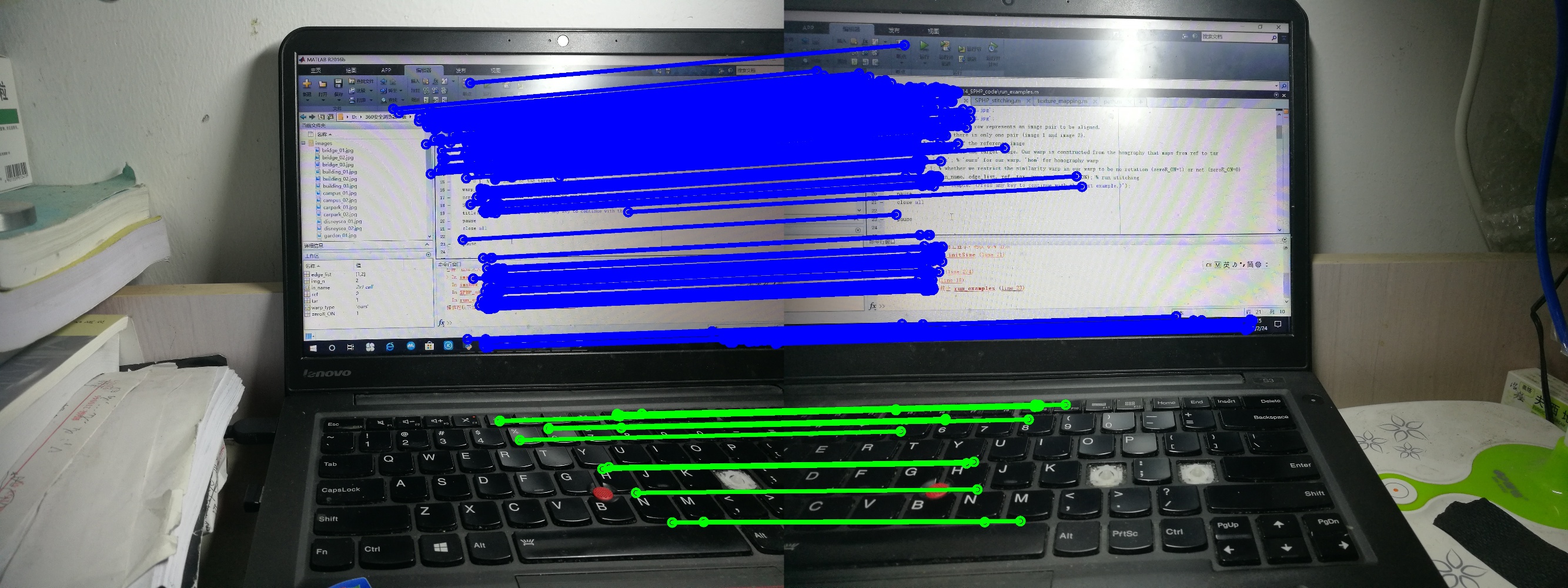}
			%\caption{fig2}
		\end{minipage}
	}%
	\subfigure[\scriptsize APAP]{
		\begin{minipage}{0.32\textwidth}
			\centering
			\includegraphics[width=1.0in]{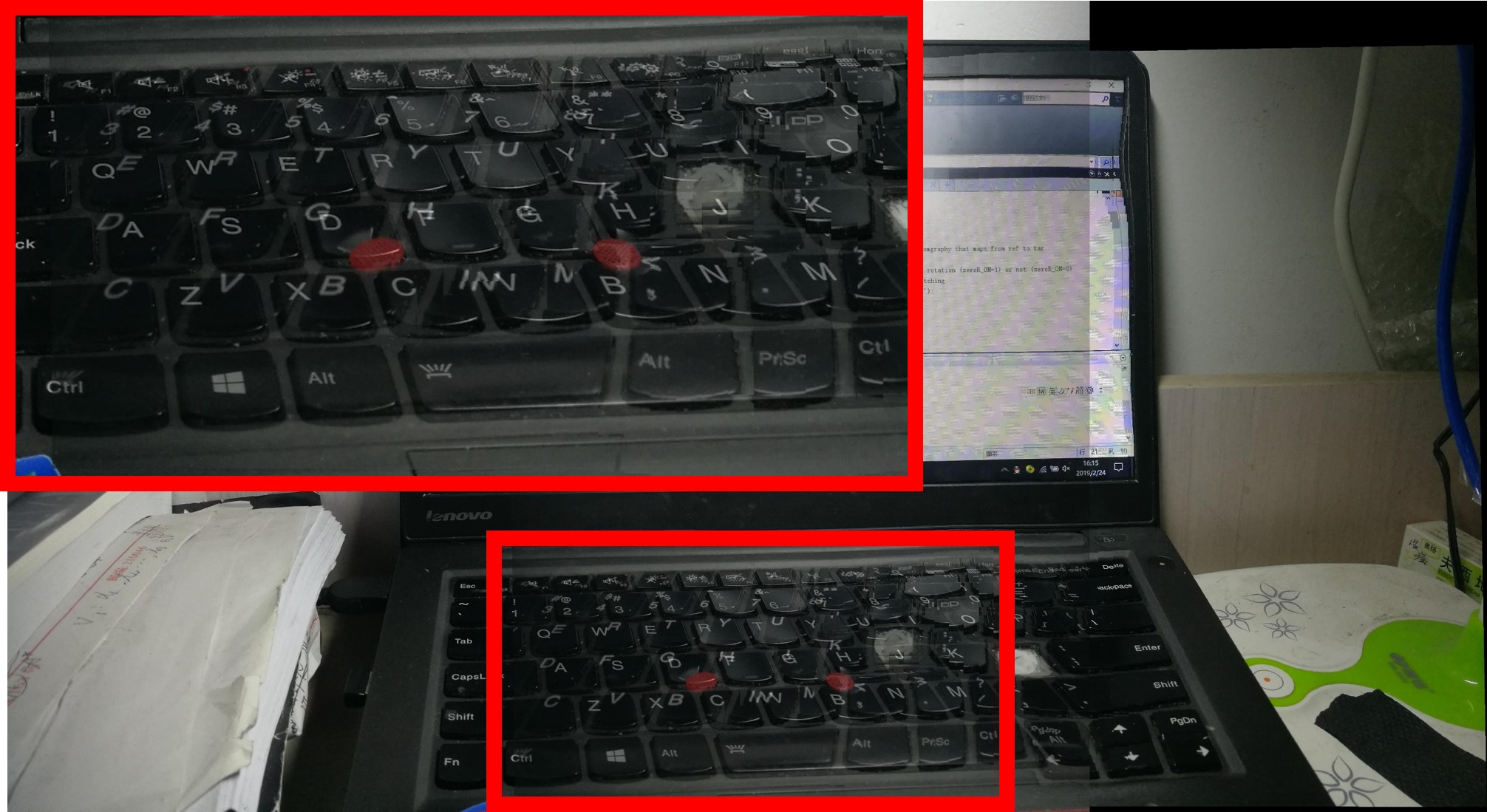}
			%\caption{fig2}
		\end{minipage}
	}%
	
	\subfigure[\scriptsize ICE]{
		\begin{minipage}{0.32\textwidth}
			\centering
			\includegraphics[width=1.0in]{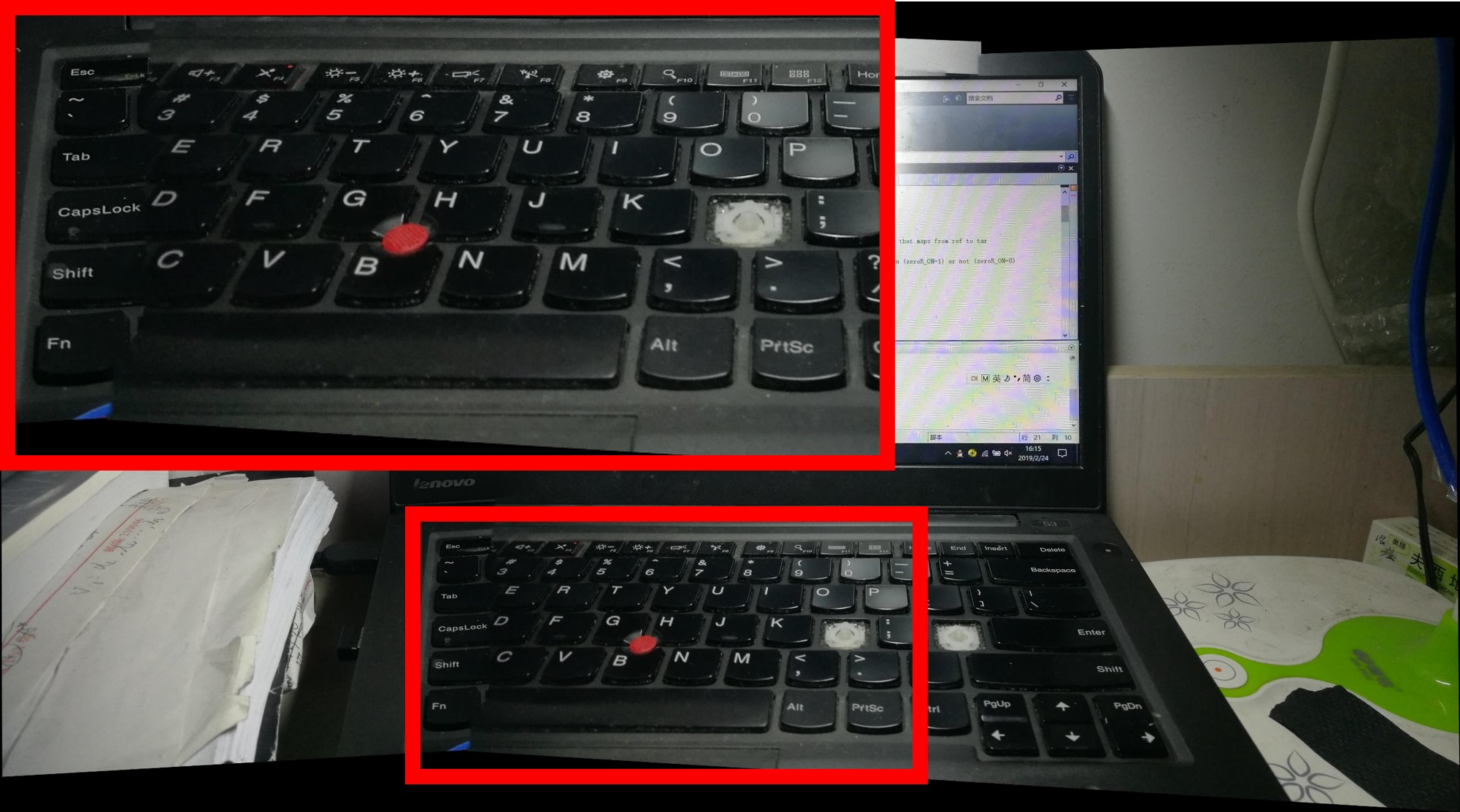}
		\end{minipage}
	}%
	\subfigure[\scriptsize NISwGSP]{
		\begin{minipage}{0.32\textwidth}
			\centering
			\includegraphics[width=1.0in]{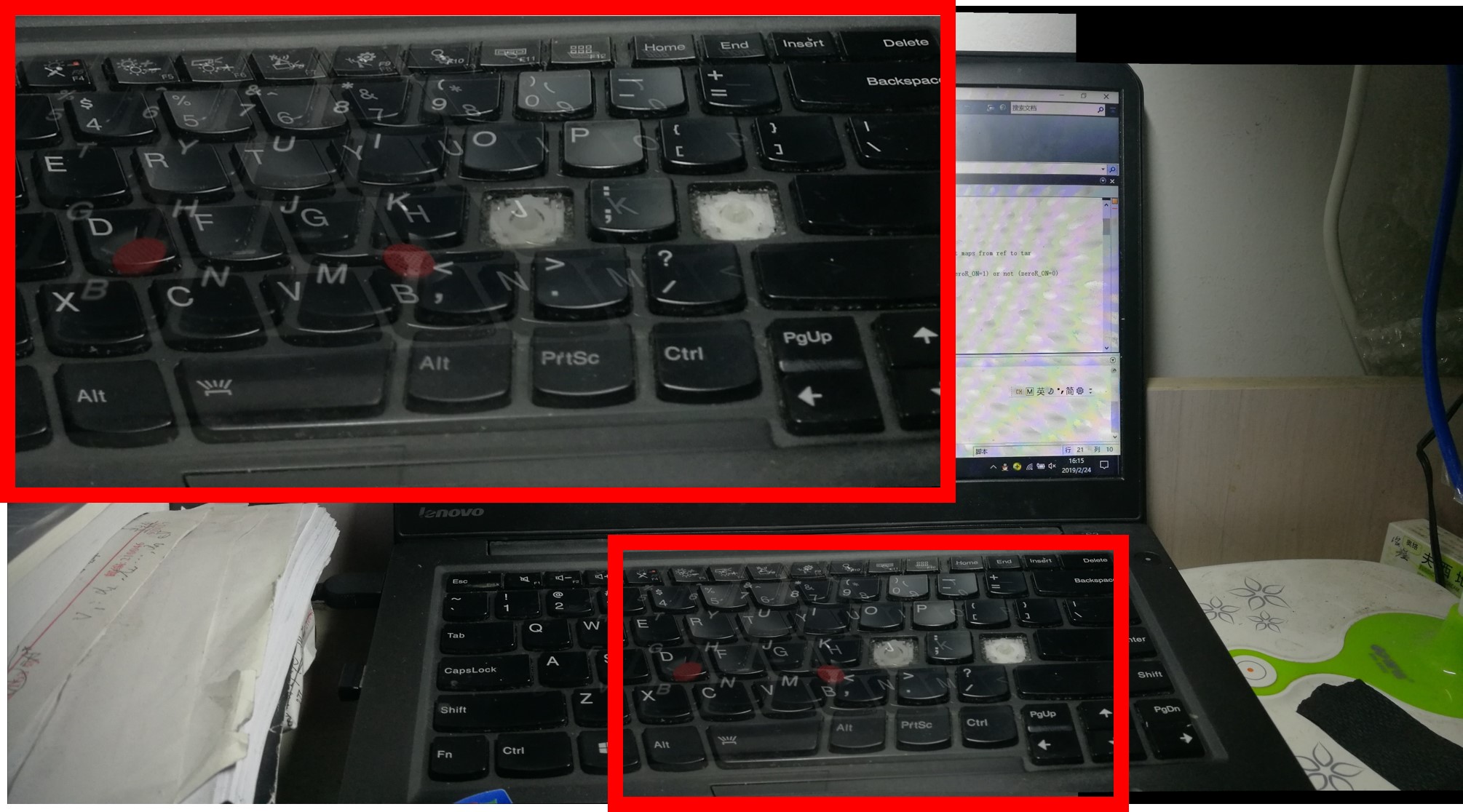}
		\end{minipage}
	}%
	\subfigure[\scriptsize Ours]{
		\begin{minipage}{0.32\textwidth}
			\centering
			\includegraphics[width=1.0in]{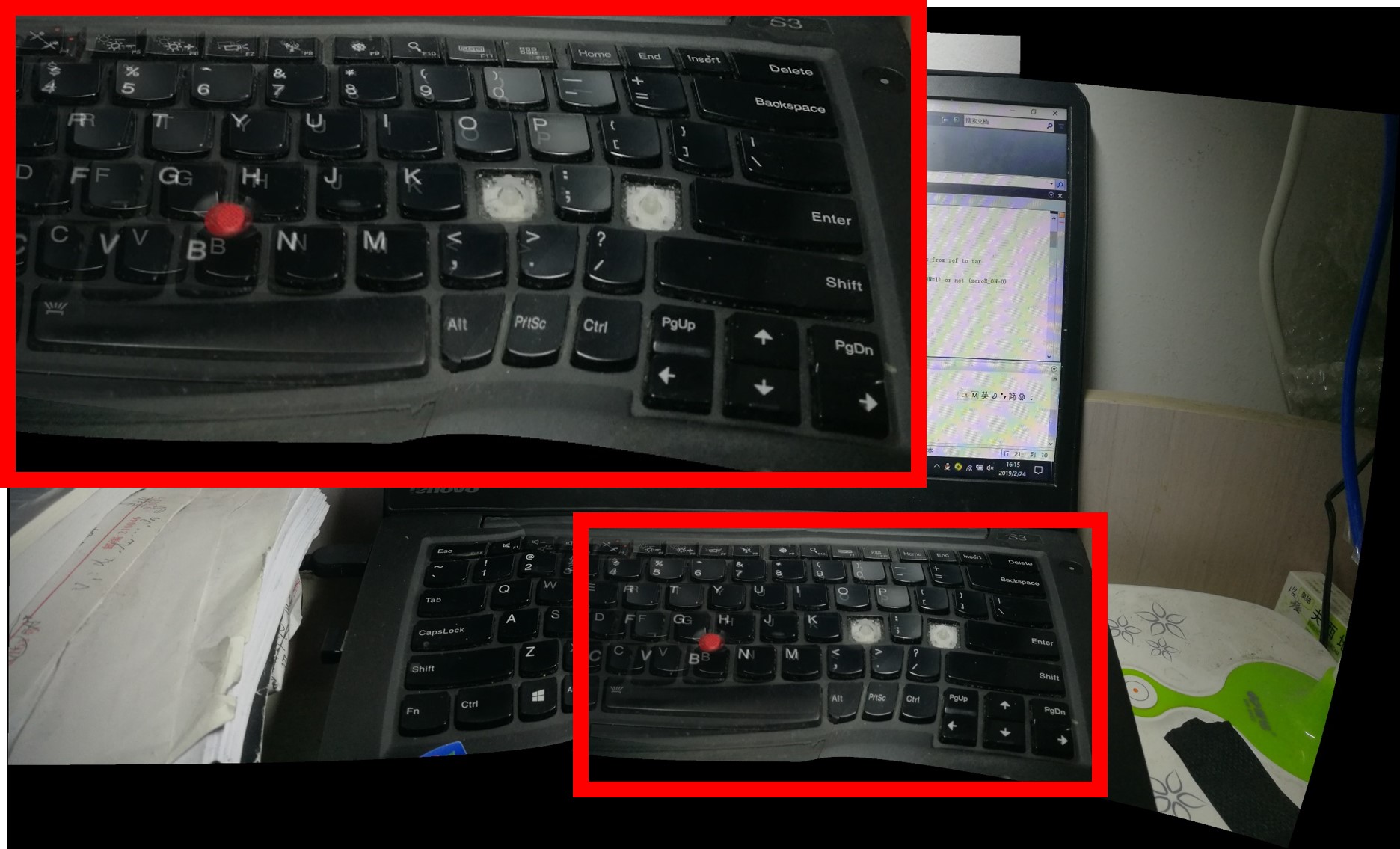}
		\end{minipage}
	}%
	
	\caption{An indoor scene taken by ourselves. The global RANSAC fails to extract matching relations on the keyboard, since the distribution of matched keypoints on it differs greatly from the screen. Though our method still shows unremarkable artifact on the keyboard, it aligns more keys correctly than other methods.}
	\label{prac0Analysis}
\end{figure}

\begin{figure}
	\centering
	\subfigure[\scriptsize Input images]{
		\begin{minipage}{0.16\textwidth}
			\centering
			\includegraphics[width=1.0\textwidth]{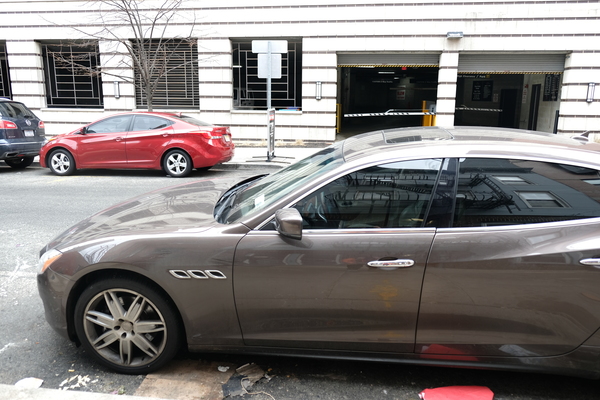}
			%\caption{fig2}
		\end{minipage}
		\begin{minipage}{0.16\textwidth}
			\centering
			\includegraphics[width=1.0\textwidth]{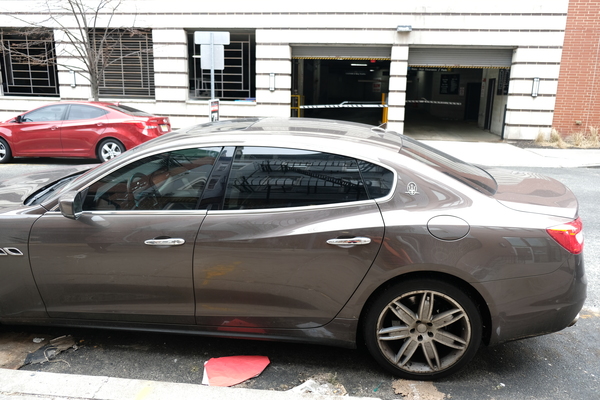}
			%\caption{fig2}
		\end{minipage}	
	}%
	\subfigure[\scriptsize RISwMR]{
		\begin{minipage}{0.32\textwidth}
			\centering
			\includegraphics[width=0.9in]{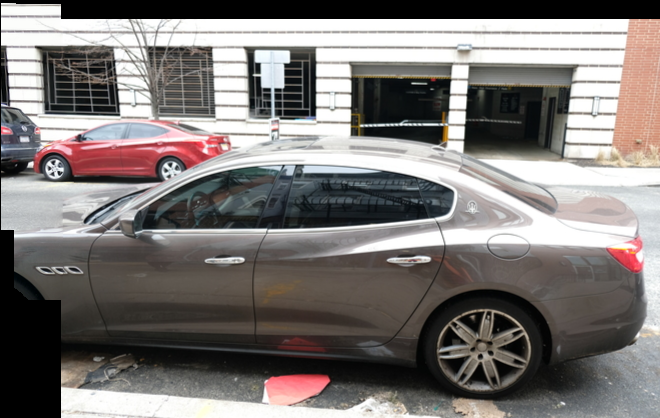}
		\end{minipage}
	}%
	\subfigure[\scriptsize Ours]{
		\begin{minipage}{0.32\textwidth}
			\centering
			\includegraphics[width=1.0in]{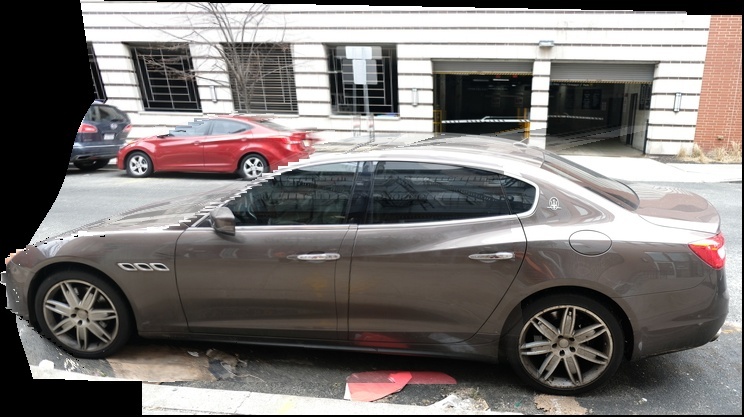}
		\end{minipage}
	}%
	
	\caption{Comparison with other multiple homography method on Car image pair (from \cite{eccv18}). With proper segmentation and regional constraint, our method reserves important parts and maintains smoothness of the final stitching result.}
	\label{carAnalysis}
\end{figure}

\begin{table}
	\centering
	\caption{Quantitative comparisons measured by the RMSE-NCC score on our evaluation dataset. For each image pair, the best is shown in bold. Our method achieves the best performance over the whole dataset.}
	\begin{tabular}{|c|c|c|c|c|c|c|c|c|c|c|}
		\hline
		\textbf{No.} & \textbf{APAP} & \textbf{SPHP} & \textbf{NISwGSP} & \textbf{PRCS} &  &  \textbf{No.} & \textbf{APAP} & \textbf{SPHP} & \textbf{NISwGSP} & \textbf{PRCS} \\
		\hline
		\hline
		\textbf{01} &13.289 &16.119 &11.887 &\textbf{11.108} & &\textbf{11} &7.152 &7.218 &\textbf{6.508} &6.722 \\
		\hline
		\textbf{02} &\textbf{4.457} &4.854 &5.043 &4.583 & &\textbf{12} &\textbf{1.031} &1.709 &1.794 &1.822 \\
		\hline
		\textbf{03} &6.456 &8.360 &6.510 &\textbf{6.395} & &\textbf{13} &9.273 &9.218 &7.791 &\textbf{7.627} \\
		\hline
		\textbf{04} &15.587 &18.448 &\textbf{10.464} &11.022 & &\textbf{14} &6.715 &6.502 &\textbf{5.711} &\textbf{5.711} \\
		\hline
		\textbf{05} &6.865 &9.301 &\textbf{6.817} &6.935 & &\textbf{15} &4.155 &4.803 &\textbf{3.060} &3.952 \\
		\hline
		\textbf{06} &7.001 &7.303 &6.771 &\textbf{6.144} & &\textbf{16} &5.389 &5.707 &\textbf{5.325} &5.369 \\
		\hline
		\textbf{07} &14.040 &15.720 &11.906 &\textbf{9.863} & &\textbf{17} &\textbf{1.123} &2.112 &2.045 &2.083 \\
		\hline
		\textbf{08} &14.052 &11.698 &12.750 &\textbf{11.040} & &\textbf{18} &15.370 &14.372 &13.856 &\textbf{13.791} \\
		\hline
		\textbf{09} &9.206 &8.923 &7.869 &\textbf{5.562} & &\textbf{19} &6.455 &6.339 &\textbf{5.694} &5.746 \\
		\hline
		\textbf{10} &9.271 &9.323 &\textbf{7.538} &7.607 & &\textbf{20} &9.670 &9.293 &\textbf{8.829} &8.858 \\
		\hline
		\hline
		\multicolumn{2}{|c|}{\multirow{2}*{Average Score}} & \multicolumn{2}{|c|}{\textbf{APAP}}
		& \multicolumn{3}{|c|}{\textbf{SPHP}}
		&\multicolumn{2}{|c|}{\textbf{NISwGSP}}
		&\multicolumn{2}{|c|}{\textbf{PRCS}} \\
		\cline{3-11}
		\multicolumn{2}{|c|}{~} &
		\multicolumn{2}{|c|}{8.328} & \multicolumn{3}{|c|}{8.866} &
		\multicolumn{2}{|c|}{7.408} &
		\multicolumn{2}{|c|}{\textbf{7.097}} \\
		\hline
	\end{tabular}
	\label{results}
\end{table}

For APAP, SPHP and NISwGSP, we tune the parameters according to the guideline suggested by the authors to achieve the best results. The RMSE-NCC scores are shown in Table \ref{results}. As we can see, our method generally shows superior performance over APAP and SPHP and similar performance with NISwGSP in most cases. However, in scenes where the parallax is big (like 7 and 9 from our evaluation dataset), our method exhibits superior performance over NISwGSP due to proper regional constraints and transformations. Please note that due to space limit, the comparisons using PSNR and SSIM as the measures are included in our accompanying material. 

\subsection{Qualitative Analysis}

As a core part of our method, planar region consensus and relevant constraints contribute greatly to high quality stitching in challenging scenes.

\begin{figure}
	\centering
	\subfigure[\scriptsize RANSAC comparison]{
	\begin{minipage}{0.33\textwidth}
		\centering
		\includegraphics[width=1.0in]{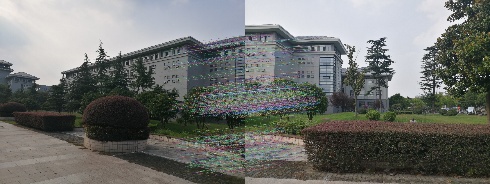}\vspace{4pt}
		\includegraphics[width=1.0in]{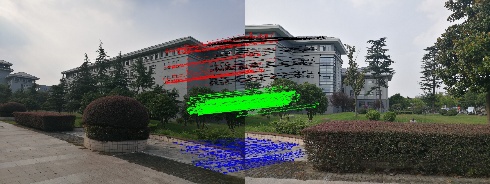}
		%\caption{fig2}
	\end{minipage}
	}%
	\subfigure[\scriptsize APAP]{
		\begin{minipage}{0.33\textwidth}
			\centering
			\includegraphics[width=0.9in]{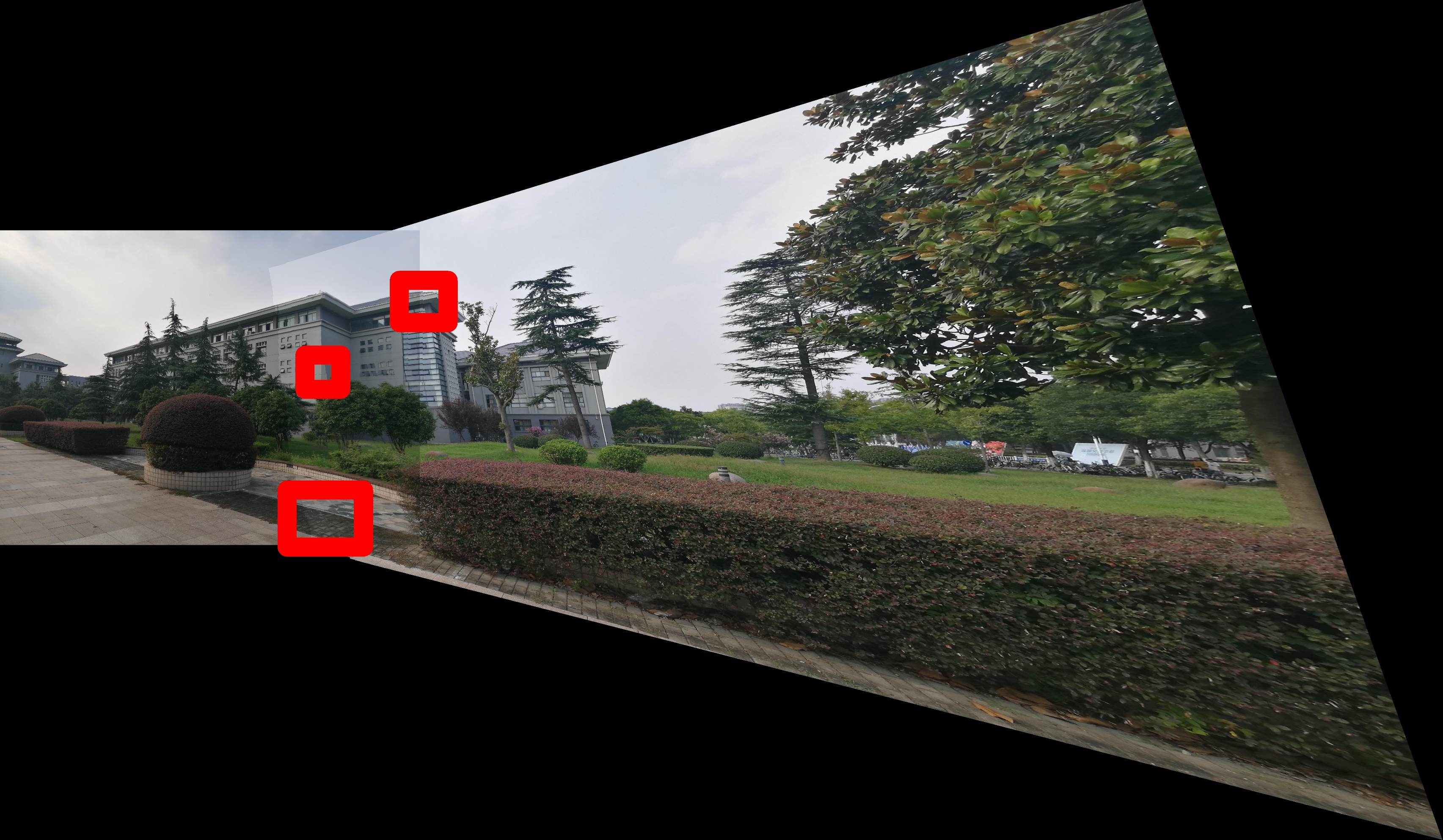}\vspace{4pt}
			\includegraphics[width=1.2in]{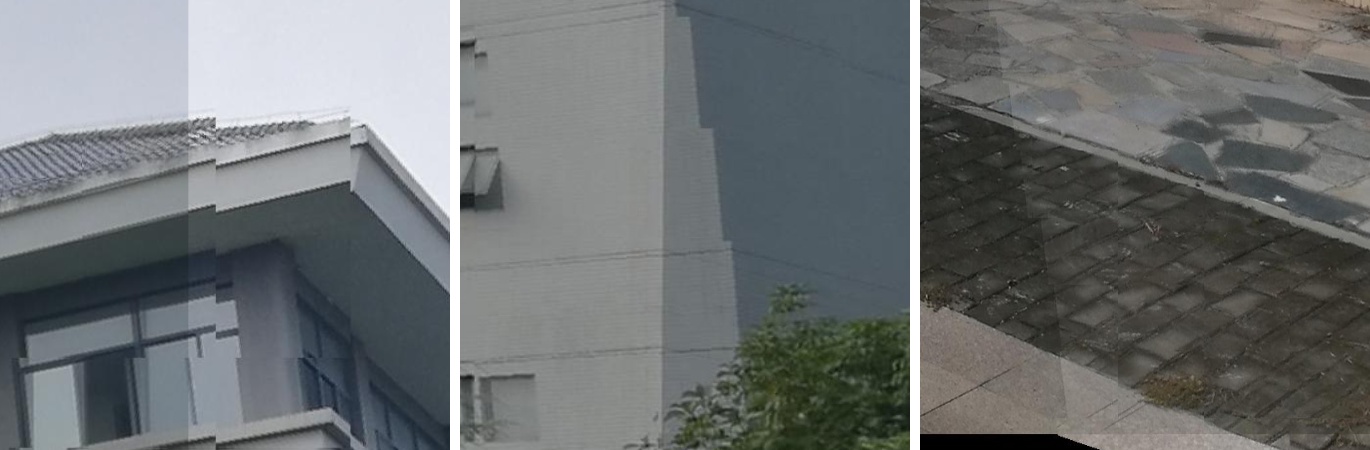}
			%\caption{fig2}
		\end{minipage}
	}%
	\subfigure[\scriptsize SPHP]{
		\begin{minipage}{0.33\textwidth}
			\centering
			\includegraphics[width=1.2in]{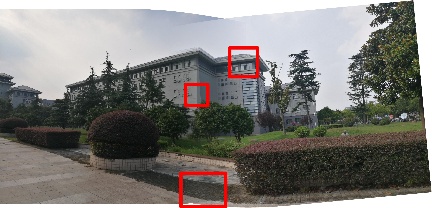}\vspace{4pt}
			\includegraphics[width=1.2in]{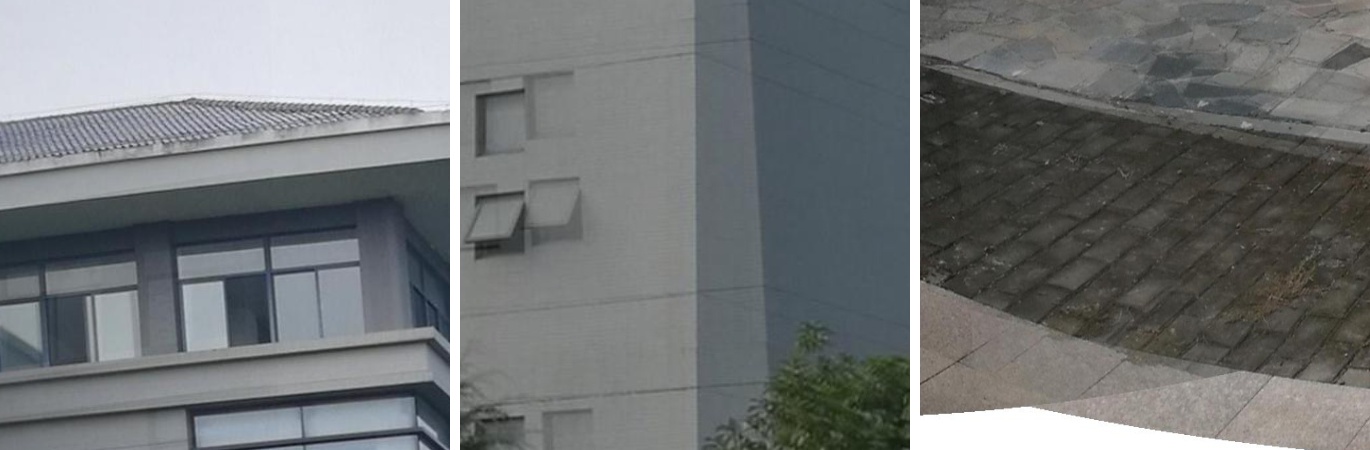}
			%\caption{fig2}
		\end{minipage}
	}%

	\subfigure[\scriptsize NISwGSP]{
		\begin{minipage}{0.33\textwidth}
			\centering
			\includegraphics[width=1.2in]{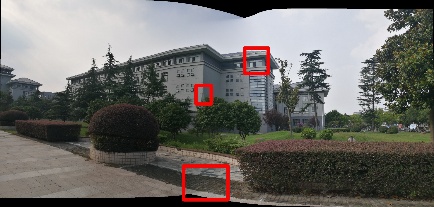}\vspace{4pt}
			\includegraphics[width=1.2in]{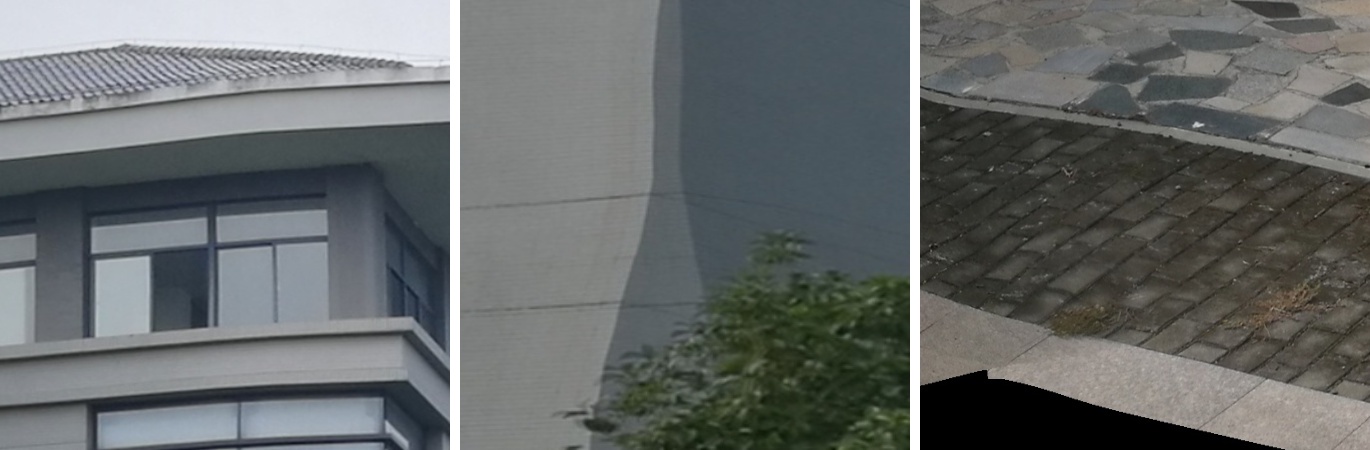}
			%\caption{fig2}
		\end{minipage}
	}%
	\subfigure[\scriptsize ICE]{
		\begin{minipage}{0.33\textwidth}
			\centering
			\includegraphics[width=1.2in]{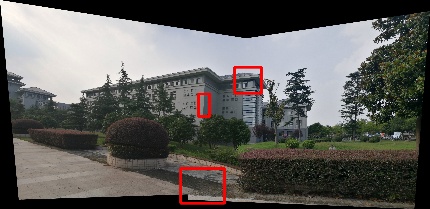}\vspace{4pt}
			\includegraphics[width=1.2in]{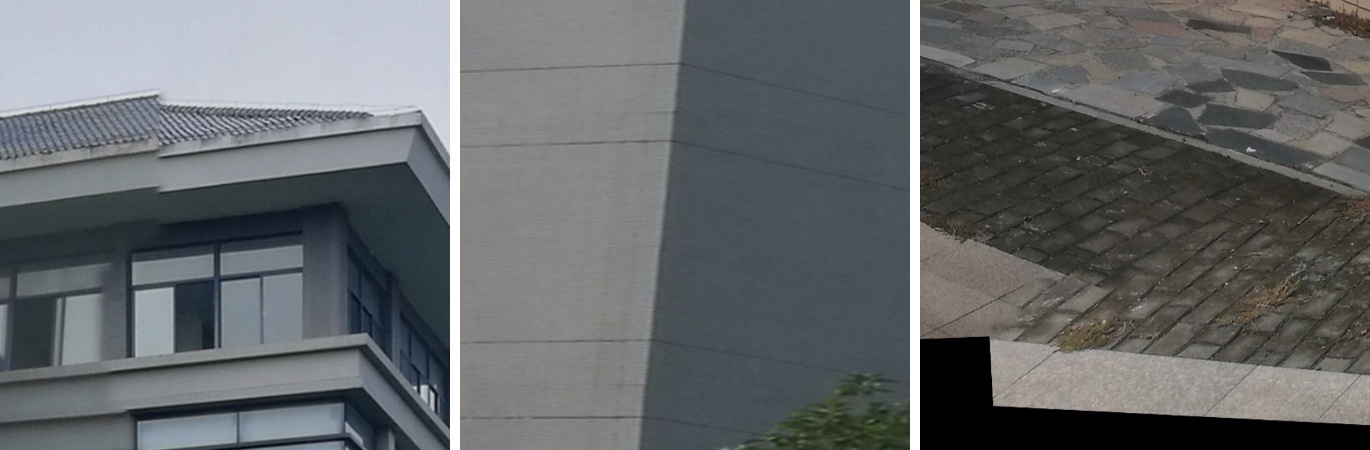}
			%\caption{afig2}
		\end{minipage}
	}%
	\subfigure[\scriptsize Ours]{
		\begin{minipage}{0.33\textwidth}
			\centering
			\includegraphics[width=1.2in]{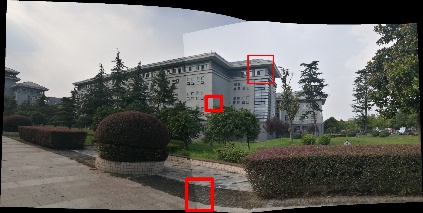}\vspace{4pt}
			\includegraphics[width=1.2in]{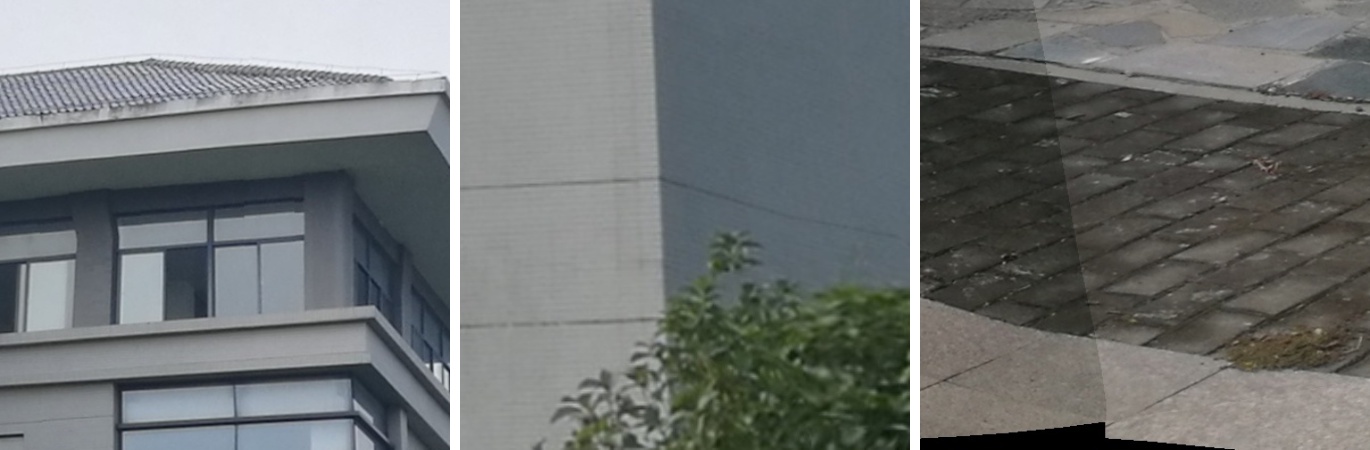}
			%\caption{fig2}
		\end{minipage}
	}%
	
	\caption{Qualitative analysis on image pair 10 from the evaluation dataset. Though the global and regional RANSACs find almost the same matched keypoints, planar region knowledge helps to group them correctly. With these regional information, our method aligns images correctly in the regions highlighted.}
	\label{seu4Analysis}
\end{figure}

Figs. \ref{cornerAnalysis} and \ref{prac0Analysis} show examples in which our regional RANSACs find better keypoint matching results. The keypoint distribution on the ground plane of Fig. \ref{cornerAnalysis} varies significantly from the outer wall. The same happens in the keyboard of Fig. \ref{prac0Analysis} which hinders a global RANSAC from extracting the matched pattern. With this additional group correctly extracted, our method naturally outperforms previous ones which fail to find correct matching relations.

Fig. \ref{carAnalysis} demonstrates the superiority of our method over other state-of-the-art multiple registration methods. Though these methods extract the same number of matching groups and obtain similar local models as ours, we extend the stitching result from the overlapping area through regional transformation information. This helps to reserve image information as much as possible and avoid losing image content as RISwMR.

Fig. \ref{seu4Analysis} shows the advantage of applying planar region consensus throughout our mesh optimization framework. In this challenging image pair, multiple dominant planar regions exist in the overlapping area. Our method correctly finds out all four matched dominant regions and processes them respectively. Without correct grouping, other methods fail to compensate for all details of the image, leading to visual artifacts such as misalignments or ghostings.

\section{Conclusion}

We have presented a novel image stitching method based on planar region consensus, a kind of latent semantics which can be effectively extracted by our new scheme on building a planar region segmentation network. Using planar region knowledge, we design a method which fully exploits regional information, obtains accurate local alignments and maintains transition naturalness. Both quantitative evaluation and qualitative analysis demonstrate our superiority over the state-of-the-arts.

Apart from image stitching, we believe planar region knowledge will also benefit other vision tasks. Since all pixels lie approximately in the same planar region, they probably exhibit similar photometric properties which can be utilized in the tasks such as light estimation.

We also observe several limitations. First, the quality of matched planar regions depends on segmentation result. For an image pair whose segmentation is bad, like over-segmentation, keypoints belonging to the same region may be divided into different groups, influencing the local models. Second, in low-texture areas where no valid matching can be obtained, our method has difficulty in finding good transformation as well. We plan to investigate them in future work.

\clearpage
\bibliographystyle{splncs04}
\bibliography{egbib}

\begin{thebibliography}{10}
\providecommand{\url}[1]{\texttt{#1}}
\providecommand{\urlprefix}{URL }
\providecommand{\doi}[1]{https://doi.org/#1}

\bibitem{SPHP}
Chang, C.H., Sato, Y., Chuang, Y.Y.: Shape-preserving half-projective warps for
  image stitching. pp. 3254--3261 (06 2014). \doi{10.1109/CVPR.2014.422}

\bibitem{NISwGSP}
Chen, Y.S., Chuang, Y.Y.: Natural image stitching with the global similarity
  prior. In: Leibe, B., Matas, J., Sebe, N., Welling, M. (eds.) Computer Vision
  -- ECCV 2016. pp. 186--201. Springer International Publishing, Cham (2016)

\bibitem{SIFT}
David, G.: Distinctive image features from scale-invariant keypoints. J.
  Comput. Vis  \textbf{147},  91--110 (01 2004)

\bibitem{efficientGraphBasedImageSegmentation}
Felzenszwalb, P., Huttenlocher, D.: Efficient graph-based image segmentation.
  International Journal of Computer Vision  \textbf{59},  167--181 (09 2004).
  \doi{10.1023/B:VISI.0000022288.19776.77}

\bibitem{RANSACOrig}
Fishler, M.: Bolles: Random sample consensus: A paradigm for model fitting with
  applications to image analysis and automated cartography. Communications of
  The ACM - CACM  (01 1981)

\bibitem{dualHomographyWarp}
{Gao}, J., {Kim}, S.J., {Brown}, M.S.: Constructing image panoramas using
  dual-homography warping. In: CVPR 2011. pp. 49--56 (June 2011).
  \doi{10.1109/CVPR.2011.5995433}

\bibitem{seamDrivenImageStitching}
Gao, J., Yu, L., Chin, T.J., Brown, M.: Seam-driven image stitching (05 2013)

\bibitem{LSD:ALineSegmentDetector}
Gioi, R., Jakubowicz, J., Morel, J.M., Randall, G.: Lsd: A line segment
  detector. Image Processing On Line  \textbf{2},  35--55 (03 2012).
  \doi{10.5201/ipol.2012.gjmr-lsd}

\bibitem{parallaxRobustSurveillanceVideoStitching}
He, B., Yu, S.: Parallax-robust surveillance video stitching. Sensors
  \textbf{16}, ~7 (12 2015). \doi{10.3390/s16010007}

\bibitem{rectanglingPanoramicImagesViaWarping}
He, K., Chang, H., Sun, J.: Rectangling panoramic images via warping. ACM
  Transactions on Graphics (TOG)  \textbf{32} (07 2013).
  \doi{10.1145/2461912.2462004}

\bibitem{eccv18}
Herrmann, C., Wang, C., Bowen, R.S., Keyder, E., Krainin, M., Liu, C., Zabih,
  R.: Robust image stitching with multiple registrations. In: Ferrari, V.,
  Hebert, M., Sminchisescu, C., Weiss, Y. (eds.) Computer Vision -- ECCV 2018.
  pp. 53--69. Springer International Publishing, Cham (2018)

\bibitem{implementingAsRigidAsPossible}
Igarashi, T., Igarashi, Y.: Implementing as-rigid-as-possible shape
  manipulation and surface flattening. Journal of Graphics, GPU, and Game Tools
   \textbf{14}(1),  17--30 (2009). \doi{10.1080/2151237X.2009.10129273},
  \url{https://doi.org/10.1080/2151237X.2009.10129273}

\bibitem{imageStitchUzMultiHomoEstimatBySegRegForDiffParallax}
Lee, D., Yoon, J., Lim, S.: Image stitching using multiple homographies
  estimated by segmented regions for different parallaxes. pp. 71--75 (09
  2017). \doi{10.1109/ICVISP.2017.19}

\bibitem{Quasi-homographyWarpInImageStitching}
Li, N., Xu, Y., Wang, C.: Quasi-homography warps in image stitching. IEEE
  Transactions on Multimedia  \textbf{PP} (01 2017).
  \doi{10.1109/TMM.2017.2771566}

\bibitem{DualFeatureWarp}
Li, S., Yuan, L., Sun, J., Quan, L.: Dual-feature warping-based motion model
  estimation (12 2015). \doi{10.1109/ICCV.2015.487}

\bibitem{AANAP}
Lin, C.C., Pankanti, S., Natesan~Ramamurthy, K., Aravkin, A.: Adaptive
  as-natural-as-possible image stitching. pp. 1155--1163 (06 2015).
  \doi{10.1109/CVPR.2015.7298719}

\bibitem{directPhotoAlignByMeshDefo}
{Lin}, K., {Jiang}, N., {Liu}, S., {Cheong}, L., {Do}, M., {Lu}, J.: Direct
  photometric alignment by mesh deformation. In: 2017 IEEE Conference on
  Computer Vision and Pattern Recognition (CVPR). pp. 2701--2709 (July 2017).
  \doi{10.1109/CVPR.2017.289}

\bibitem{SEAGULL}
Lin, K., Jiang, N., Cheong, L.F., Do, M., Lu, J.: Seagull: Seam-guided local
  alignment for parallax-tolerant image stitching. In: Leibe, B., Matas, J.,
  Sebe, N., Welling, M. (eds.) Computer Vision -- ECCV 2016. pp. 370--385.
  Springer International Publishing, Cham (2016)

\bibitem{smoothlyVaryingAffineStitching}
Lin, w.y., Liu, S., Matsushita, Y., Ng, T.T., Cheong, L.: Smoothly varying
  affine stitching. pp. 345--352 (06 2011). \doi{10.1109/CVPR.2011.5995314}

\bibitem{PlaneNet}
{Liu}, C., {Yang}, J., {Ceylan}, D., {Yumer}, E., {Furukawa}, Y.: Planenet:
  Piece-wise planar reconstruction from a single rgb image. In: 2018 IEEE/CVF
  Conference on Computer Vision and Pattern Recognition. pp. 2579--2588 (June
  2018). \doi{10.1109/CVPR.2018.00273}

\bibitem{PlaneRCNN}
Liu, C., Kim, K., Gu, J., Furukawa, Y., Kautz, J.: Planercnn: 3d plane
  detection and reconstruction from a single image (2018)

\bibitem{ImageAlignmentByPiecePlanarRegionMatching}
{Lou}, Z., {Gevers}, T.: Image alignment by piecewise planar region matching.
  IEEE Transactions on Multimedia  \textbf{16}(7),  2052--2061 (Nov 2014).
  \doi{10.1109/TMM.2014.2346476}

\bibitem{UNET}
Ronneberger, O., Fischer, P., Brox, T.: U-net: Convolutional networks for
  biomedical image segmentation. In: Navab, N., Hornegger, J., Wells, W.M.,
  Frangi, A.F. (eds.) Medical Image Computing and Computer-Assisted
  Intervention -- MICCAI 2015. pp. 234--241. Springer International Publishing,
  Cham (2015)

\bibitem{fullyConvNetworksForSemanticSeg}
Shelhamer, E., Long, J., Darrell, T.: Fully convolutional networks for semantic
  segmentation. IEEE Transactions on Pattern Analysis and Machine Intelligence
  \textbf{39}, ~1--1 (05 2016). \doi{10.1109/TPAMI.2016.2572683}

\bibitem{ImageAlignmentNStitchingTutorial}
Szeliski, R.: Image alignment and stitching: A tutorial. Foundations and Trends
  in Computer Graphics and Vision  \textbf{2} (01 2006).
  \doi{10.1561/0600000009}

\bibitem{UperNet101}
Xiao, T., Liu, Y., Zhou, B., Jiang, Y., Sun, J.: Unified perceptual parsing for
  scene understanding. In: Ferrari, V., Hebert, M., Sminchisescu, C., Weiss, Y.
  (eds.) Computer Vision -- ECCV 2018. pp. 432--448. Springer International
  Publishing, Cham (2018)

\bibitem{APAP}
{Zaragoza}, J., {Chin}, T., {Tran}, Q., {Brown}, M.S., {Suter}, D.:
  As-projective-as-possible image stitching with moving dlt. IEEE Transactions
  on Pattern Analysis and Machine Intelligence  \textbf{36}(7),  1285--1298
  (July 2014). \doi{10.1109/TPAMI.2013.247}

\bibitem{ParallaxTolerantImageStitching}
Zhang, F., Liu, F.: Parallax-tolerant image stitching. pp. 3262--3269 (06
  2014). \doi{10.1109/CVPR.2014.423}

\bibitem{PCPS}
Zheng, J., Wang, Y., Wang, H., Li, B., Hu, H.M.: A novel projective-consistent
  plane based image stitching method. IEEE Transactions on Multimedia
  \textbf{PP}, ~1--1 (03 2019). \doi{10.1109/TMM.2019.2905692}

\bibitem{zhou2017scene}
Zhou, B., Zhao, H., Puig, X., Fidler, S., Barriuso, A., Torralba, A.: Scene
  parsing through ade20k dataset. In: Proceedings of the IEEE Conference on
  Computer Vision and Pattern Recognition (2017)

\bibitem{zhou2018semanticCode}
Zhou, B., Zhao, H., Puig, X., Xiao, T., Fidler, S., Barriuso, A., Torralba, A.:
  Semantic understanding of scenes through the ade20k dataset. International
  Journal on Computer Vision  (2018)

\end{thebibliography}
\end{document}